%% file: main.tex
\documentclass[10pt,twocolumn,letterpaper]{article}

\usepackage{algorithm}
\usepackage{algpseudocode}
\usepackage{algorithmicx}
\usepackage{amsmath}
\usepackage{amssymb}
\usepackage{booktabs}
\usepackage{multirow}
\usepackage{graphicx}
\usepackage{subcaption}
\usepackage{adjustbox}

\usepackage[dvipsnames,table]{xcolor}

\usepackage{amsthm}

\usepackage[algorithms]{wacv}      


\definecolor{wacvblue}{rgb}{0.21,0.49,0.74}
\usepackage[pagebackref,breaklinks,colorlinks,allcolors=wacvblue]{hyperref}
\usepackage[table]{xcolor}

\def\wacvPaperID{1419} 
\def\confName{WACV}
\def\confYear{2026}

\title{Optimal Transport for Rectified Flow Image Editing: Unifying Inversion-Based and Direct Methods}

\author{
\begin{tabular}{cc}
Marian Lupa\c{s}cu$^1$ & Mihai Sorin Stupariu$^1$ \\
{\tt\small marianlupascu15@gmail.com} & {\tt\small stupariu@fmi.unibuc.ro} \\
\multicolumn{2}{c}{$^1$Department of Computer Science, University of Bucharest, Romania}
\end{tabular}
}

\begin{document}

\twocolumn[{%
\renewcommand\twocolumn[1][]{#1}%
\maketitle
\includegraphics[width=1\linewidth]{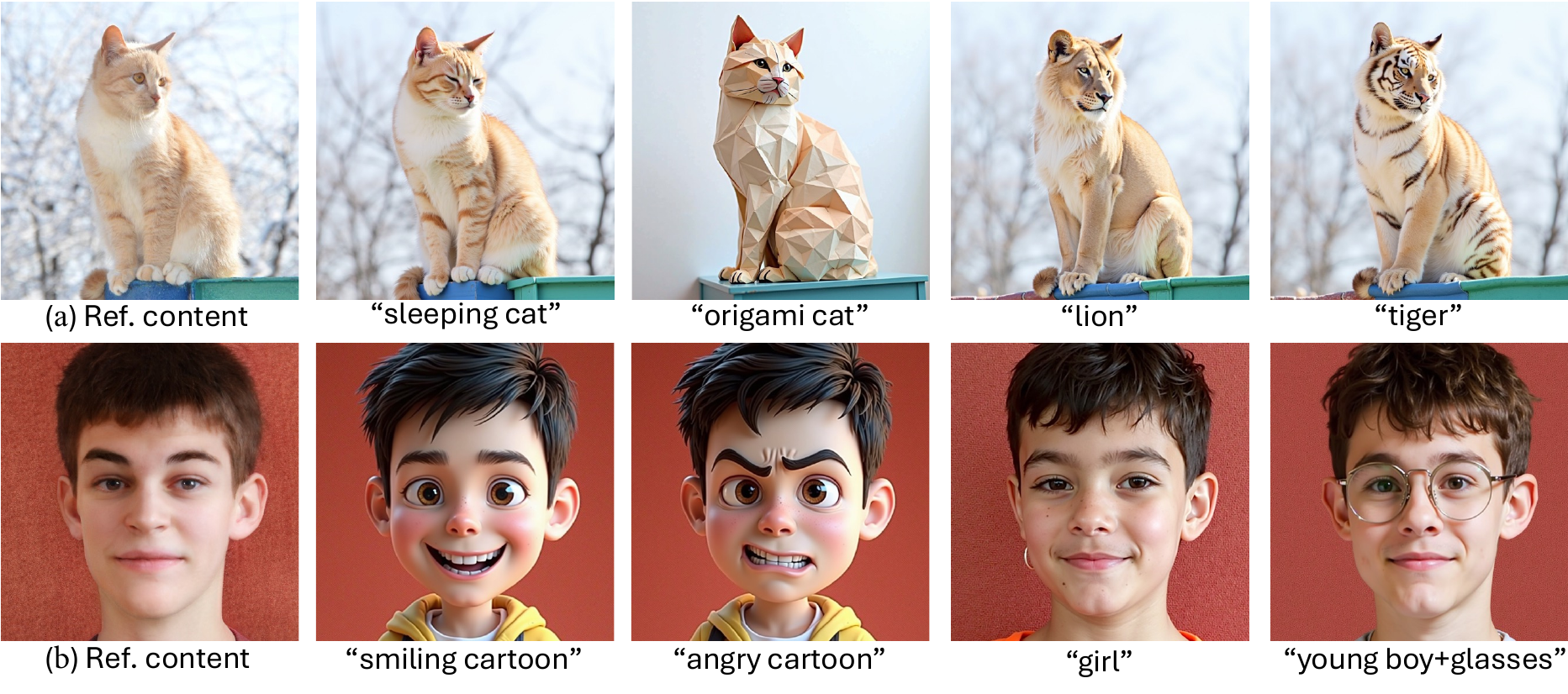}
\vspace{-2.2em}
\captionof{figure}{\textbf{Transport-guided rectified flow editing enables image modifications across both paradigms.} Our framework enhances inversion-based and inversion-free methods, applying prompt-based transformations without fine-tuning. We demonstrate cross-species edits for a cat (``lion'', ``origami cat'') and attribute edits for a face (``girl'', ``smiling cartoon''), while preserving structure.}

\vspace{1.5em}}
\label{fig:teaser}
]

\input{sec/0_abstract}    
\input{sec/1_intro}
\input{sec/2_related_works}

\input{sec/3_approach}
\input{sec/4_experiments}
\input{sec/5_conclusion}

{
    \small
    \bibliographystyle{ieeenat_fullname}
    \bibliography{main}
}

\clearpage
\appendix

\input{sec/6_appendix}

\end{document}

%% file: sec/0_abstract.tex
\begin{abstract}
Image editing in rectified flow models remains challenging due to the fundamental trade-off between reconstruction fidelity and editing flexibility. While inversion-based methods suffer from trajectory deviation, recent inversion-free approaches like FlowEdit offer direct editing pathways but can benefit from additional guidance to improve structure preservation. In this work, we demonstrate that optimal transport theory provides a unified framework for improving both paradigms in rectified flow editing. We introduce a zero-shot transport-guided inversion framework that leverages optimal transport during the reverse diffusion process, and extend optimal transport principles to enhance inversion-free methods through transport-optimized velocity field corrections. Incorporating transport-based guidance can effectively balance reconstruction accuracy and editing controllability across different rectified flow editing approaches. For inversion-based editing, our method achieves high-fidelity reconstruction with LPIPS scores of 0.001 and SSIM of 0.992 on face editing benchmarks, observing 7.8\% to 12.9\% improvements over RF-Inversion on LSUN datasets. For inversion-free editing with FlowEdit on FLUX and Stable Diffusion 3, we demonstrate consistent improvements in semantic consistency and structure preservation across diverse editing scenarios. Our semantic face editing experiments show an 11.2\% improvement in identity preservation and enhanced perceptual quality. The unified optimal transport framework produces visually compelling edits with superior detail preservation across both inversion-based and direct editing paradigms. Code is available for RF-Inversion and FlowEdit at \url{https://github.com/marianlupascu/OT-RF}
\end{abstract}

%% file: sec/1_intro.tex
\section{Introduction}

Rectified flow (RF) models~\cite{liu2022flow,esser2024scaling} achieve high-quality text-to-image generation with computational advantages over diffusion models~\cite{ho2020denoising,song2021denoising}, as demonstrated by systems like FLUX~\cite{flux} or Stable Diffusion 3~\cite{SD3}. However, effective image editing in rectified flows remains challenging, requiring either mapping real images back to editable latent representations through inversion, or developing principled direct editing approaches that bypass inversion altogether.

Image inversion in rectified flows presents unique challenges compared to diffusion models. The deterministic nature of RF trajectories eliminates stochastic regularization that aids diffusion model inversion~\cite{ho2020denoising}, leading to trajectory deviation during the inversion process. Small numerical errors compound across the deterministic flow path, causing inverted latents to drift from semantically meaningful representations~\cite{song2021score}. This creates a trade-off: methods preserving reconstruction fidelity produce latents resistant to editing, while approaches enabling modification introduce visual artifacts. Existing RF inversion approaches like DDIM inversion~\cite{song2021denoising} adapted to rectified flows accumulate discretization errors, while optimization-based methods like Null-Text Inversion~\cite{mokady2023null} require expensive test-time computation that contradicts RF efficiency advantages.

Recent work has proposed inversion-free alternatives to address these limitations. FlowEdit~\cite{FlowEdit} constructs direct pathways between source and target image distributions, bypassing the problematic inversion step entirely. While this approach achieves improved structural preservation compared to inversion-based methods, it lacks principled guidance mechanisms for trajectory optimization, potentially leading to suboptimal editing paths and semantic inconsistencies in complex transformations.

Optimal transport theory provides a mathematical framework for establishing principled pathways between probability distributions. Integration with generative models has enhanced training stability and sample quality through Wasserstein GANs~\cite{arjovsky2017wasserstein,gulrajani2017improved} and neural optimal transport~\cite{korotin2023neural}. The Wasserstein distance provides natural alignment with perceptual similarity through geometric structure preservation~\cite{peyre2019computational}, making it well-suited for semantic image editing applications across different editing paradigms.

We introduce a unified framework that leverages optimal transport theory to improve rectified flow editing across both paradigms. For inversion-based methods, we propose transport-guided trajectory optimization through principled transport corrections. For inversion-free methods, we enhance FlowEdit with optimal transport guidance for improved semantic consistency.

Our main contributions are:
\textbf{(i)} A theoretical framework linking optimal transport distances to RF editing quality across different editing approaches;
\textbf{(ii)} Transport-guided inversion with adaptive scheduling for optimal reconstruction-editability trade-offs;
\textbf{(iii)} Transport-enhanced FlowEdit with improved pathway optimization and structural preservation;
\textbf{(iv)} Comprehensive evaluation across FLUX~\cite{flux} and Stable Diffusion 3~\cite{SD3}, demonstrating consistent improvements in both paradigms;
\textbf{(v)} Training-free methods preserving rectified flow computational efficiency~\cite{liu2024instaflow,liu2022flow}.

%% file: sec/2_related_works.tex
\begin{figure*}[!t]
\centering
\includegraphics[width=1.0\textwidth]{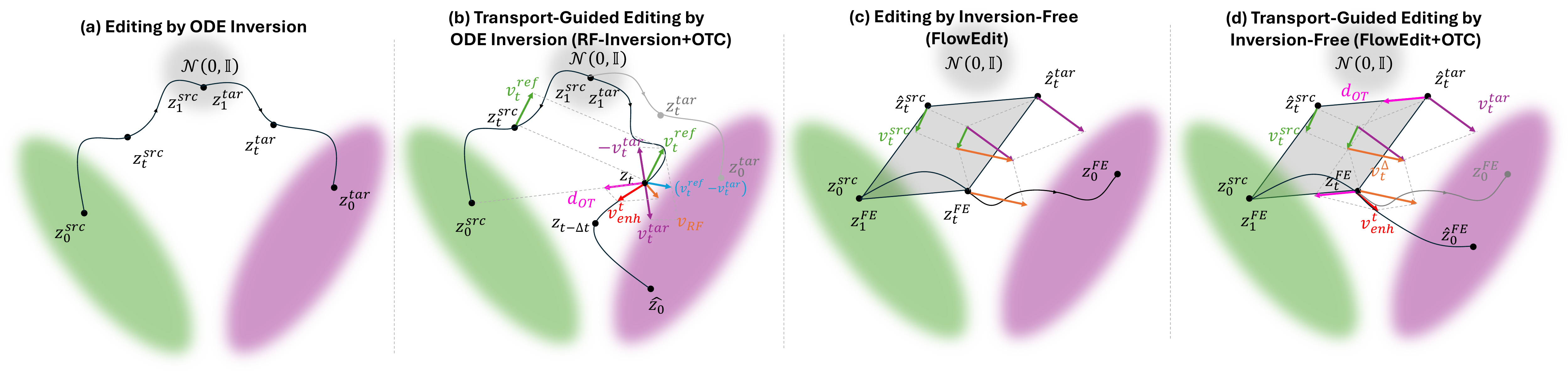}
\vspace{-0.7cm}
\caption{\textbf{Unified optimal transport framework for rectified flow editing.} (a) Standard RF inversion maps source image $z_0^{\text{src}}$ to noise, then generates $z_0^{\text{tar}}$ via reverse denoising, suffering from trajectory deviation that causes drift from source manifold (green) toward target manifold (purple). (b) Transport-guided inversion (RF-Inversion+OTC) incorporates transport directions $d_{\text{OT}}$ (cyan arrows) and reference velocities $v_t^{\text{ref}}$ as trajectory constraints, ensuring denoising paths follow transport geodesics. (c) FlowEdit constructs direct pathways $z_t^{\text{FE}}$ between distributions without inversion but lacks principled trajectory optimization. (d) Transport-enhanced FlowEdit (FlowEdit+OTC) augments velocity fields with transport guidance $d_{\text{OT}}$ (magenta arrows), creating enhanced velocities $v_{\text{enh}}$ with pathway optimization. Both enhanced methods produce $\hat{z}_0$ closer to the source manifold structure while achieving target semantic alignment, demonstrating superior structure preservation compared to baselines. OTC denotes Optimal Transport Coupling. Visualization approach inspired from~\cite{FlowEdit}.}
\label{fig:intuition}
\vspace{-0.5cm}
\end{figure*}

\section{Related Work}

\textbf{Inversion-Based Image Editing.} 
Early inversion methods like DDIM (Denoising Diffusion Implicit Models) inversion~\cite{song2021denoising} encounter linear approximation errors and reconstruction challenges. DDPM Inversion~\cite{edict} achieves improved reconstruction through exact inversion but with increased computational cost. Null-text inversion (NTI)~\cite{mokady2023null} combines pivotal inversion with test-time optimization to improve reconstruction accuracy, often paired with Prompt-to-prompt (P2P)~\cite{hertz2022prompt} for cross-attention control during editing. However, NTI requires expensive optimization procedures that contradict efficiency goals.

\textbf{Training-Free Non-Attention Methods.} 
SDEdit~\cite{meng2022sdedit} approaches editing through noise addition and denoising but suffers from limited control over semantic modifications. LEDITS++~\cite{LEDITS} utilizes perfect inversion with DPM-Solver++ and implicit masking for semantic grounding, achieving efficient editing without requiring training or optimization while supporting multiple simultaneous edits. Direct Inversion~\cite{ju2024direct} introduces disentangled branches for better editability-fidelity trade-offs. These methods avoid the computational overhead of attention manipulation while maintaining editing flexibility.

\textbf{Attention-Based Editing Methods.} 
Foundational attention-based methods manipulate attention mechanisms for precise semantic control. Prompt-to-Prompt (P2P)~\cite{hertz2022prompt} enables cross-attention control by injecting attention maps during diffusion, allowing token-level editing without pixel-space specifications. MasaCtrl~\cite{masactrl2023} performs tuning-free mutual self-attention control for consistent image synthesis and editing. Plug-and-Play (PnP)~\cite{tumanyan2023plug} features spatially-aware attention control, while StyleAligned~\cite{stylealigned2023} ensures style consistency across generated image sets through shared attention mechanisms. These methods achieve fine-grained semantic control but often require complex attention processors and may not transfer seamlessly across different model architectures.

\textbf{Training-Based Methods.} 
Some approaches require model retraining or fine-tuning for editing capabilities. InstructPix2Pix~\cite{InstructPix2Pix} learns to follow text-based editing instructions through supervised training on image-edit pairs, achieving strong semantic control but requiring substantial computational resources. While effective, these approaches lack the flexibility and efficiency of training-free methods for diverse editing scenarios.

\textbf{Rectified Flows and Flow-Based Models.} 
Rectified flows~\cite{liu2022flow} introduced neural ODEs with straight-line trajectories and reflow procedures for single-step generation. Flow Matching~\cite{lipman2023flow} provided simulation-free training for continuous normalizing flows with optimal transport interpolation, while subsequent work includes InstaFlow~\cite{liu2024instaflow} for one-step text-to-image generation and scaling approaches~\cite{esser2024scaling} for high-resolution synthesis. Recent developments~\cite{wang2024taming, deng2024fireflow} explore controlled ODEs, though principled frameworks for optimal trajectory optimization remain an active research area.

\textbf{Flow-Based Image Editing.} 
Recent work has proposed flow-specific editing methods that leverage the unique properties of rectified flows. For rectified flows specifically, RF-Inversion~\cite{rout2024semantic} employs LQR (linear–quadratic regulator) formulations for optimal control, but still faces trajectory deviation challenges inherent to deterministic flow paths. FlowEdit~\cite{kulikov2024flowedit} constructs direct pathways between source and target image distributions, achieving improved structural preservation compared to inversion-based methods through lower transport costs. The approach FlowChef~\cite{patel2024flowchef} proposes vector field steering for controlled rectified flow generation, addressing inverse problems and editing through gradient-free trajectory modification. While FlowChef demonstrates effective control mechanisms, our approach differs by leveraging optimal transport theory to provide mathematically principled guidance, specifically targeting inversion quality and trajectory optimization rather than general controlled generation. While these methods demonstrate effective control mechanisms, they lack principled theoretical foundations for trajectory optimization, potentially leading to suboptimal editing paths in complex transformations.

\textbf{Optimal Transport in Generative Models.} 
Optimal transport (OT) integration has enhanced training stability and sample quality in generative models. Wasserstein GANs~\cite{arjovsky2017wasserstein,gulrajani2017improved} demonstrated OT's utility in adversarial training, while neural optimal transport~\cite{korotin2023neural} enabled scalable computation using neural networks as transport plan approximators. The Wasserstein distance aligns with perceptual similarity through geometric structure preservation~\cite{peyre2019computational}, making it well-suited for semantic image editing applications. Our work leverages optimal transport theory to provide principled guidance for both inversion-based and inversion-free rectified flow editing approaches.

%% file: sec/3_approach.tex
\section{Approach}
\label{sec:approach}
Our key insight is that optimal transport theory provides a unified mathematical framework for both inversion-based and inversion-free rectified flow editing without requiring model retraining. We propose a comprehensive approach that leverages transport-optimal trajectories: the Optimal Transport Inversion Pipeline for inversion-based editing, and Transport-Enhanced FlowEdit for direct editing approaches. Figure~\ref{fig:intuition} illustrates our unified framework, contrasting standard methods with our transport-enhanced variants across both editing paradigms.


\subsection{Overview}

Modern rectified flow (RF) models like FLUX~\citep{flux} establish deterministic mappings that transform pure Gaussian noise $z_T \sim \mathcal{N}(0,I)$ into data samples $z_0$ via straight-line trajectories in the latent space. While this deterministic framework offers computational advantages, both inversion-based and direct editing approaches face trajectory optimization challenges. Inversion methods suffer from trajectory deviation when recovering noise initialization $z_T$ from a given sample $z_0$, while direct methods like FlowEdit~\citep{kulikov2024flowedit} lack systematic optimization for pathway construction between source and target distributions.

We address these limitations by incorporating optimal transport (OT) theory~\citep{villani2003topics, arjovsky2017wasserstein} to compute minimal-cost transformations between probability distributions. For inversion-based editing, our method ensures that inversion trajectories approximate geodesics in transport space, preserving structural coherence and semantic relationships. For direct editing, we enhance FlowEdit's pathway construction with transport-theoretic optimization to improve semantic consistency and reduce trajectory deviation. Both approaches maintain the computational efficiency of rectified flows while achieving mathematical control.

\subsection{Motivation: Why Optimal Transport?}

Both inversion-based and direct rectified flow editing methods face similar fundamental challenges. Recent theoretical analysis~\citep{hertrich2024relation,lipman2023flow} reveals that rectified flows do not inherently follow optimal transport paths. For inversion methods, the straight trajectories that make RFs efficient for generation are problematic during inversion, where semantic structure preservation is crucial. For direct methods like FlowEdit, while the approach constructs shorter paths between source and target distributions, these paths lack theoretical foundation, potentially leading to suboptimal trajectories in complex transformations~\cite{FlowEdit}.

In practice, both paradigms manifest trajectory deviation—small errors compound over the denoising process, leading to semantic drift and structural inconsistencies~\citep{rout2024semantic,wang2024taming}. Optimal transport provides the missing theoretical grounding for both approaches. By computing transport maps between distributions, we obtain globally optimal paths that minimize displacement while preserving measure~\citep{villani2008optimal}, ensuring transformations follow efficient paths in the latent space while maintaining structural coherence.

\subsection{Efficient Transport Direction Computation}

Exact optimal transport computation scales poorly with dimensionality, requiring expensive iterative solvers~\citep{cuturi2013sinkhorn}. However, we can avoid this computational bottleneck by exploiting the Brenier map property~\citep{brenier1991polar}: for quadratic costs between Gaussian-like distributions, optimal transport directions admit closed-form solutions. This assumption is well-justified for rectified flow latent spaces, which exhibit approximately Gaussian structure due to the diffusion process initialization and the central limit behavior of high-dimensional embeddings.

 We define $d_{\text{OT}}$ as:
 \begin{equation}
    d_{\text{OT}}(z_t, z_{\text{target}}, t) = \frac{z_{\text{target}} - z_t}{\max(T-t, \delta)}
\end{equation}
\noindent
where $\delta = 0.01$ prevents  numerical instabilities near the endpoint $t = T$. We use Wasserstein-2 distances as they naturally align with the quadratic cost structure of rectified flows and provide closed-form solutions for Gaussian-like latent distributions. For Wasserstein-2 distances with quadratic cost $c(z_t,z_{\text{target}}) = \|z_t-z_{\text{target}}\|^2$, the gradient of the transport functional is exactly the displacement vector $z_{\text{target}} - z_t$~\citep{brenier1991polar,villani2003topics}. This eliminates iterative optimization, reducing computational complexity from $O(n^3)$ for general optimal transport to $O(n)$ for our closed-form approach.

The temporal scaling $1/(T-t)$ serves two purposes: it intensifies transport guidance as we approach the denoising endpoint where reconstruction fidelity is most critical, and it maintains consistency with geodesic flow theory in Wasserstein space~\citep{ambrosio2005gradient}. For numerical stability, we apply gradient clipping $\|d_{\text{OT}}\|_2 \leq \tau$ with $\tau = 10.0$, chosen to prevent transport directions from overwhelming the rectified flow velocity while preserving meaningful optimization.

\subsection{Transport-Guided Inversion}

The denoising process exhibits different requirements across timesteps. Early stages ($t \approx T$) prioritize global structure preservation during noise removal, while later stages ($t \approx 0$) require precise semantic alignment with prompt conditioning. We model this temporal evolution through an adaptive weighting schedule: $\alpha(t) = \beta_0 \cdot \mathcal{S}\left(\frac{T-t}{T}\right)$, where $\mathcal{S}$ implements cosine annealing: $\mathcal{S}(s) = \frac{1}{2}\left(1 + \cos\left(\min\left(\frac{s}{\phi}, 1\right) \cdot \pi\right)\right)
$. Cosine annealing~\cite{loshchilov2016sgdr} provides several advantages over linear or exponential schedules. Its smooth, differentiable profile prevents sudden changes in transport guidance that could destabilize the denoising trajectory. The convex-concave shape naturally concentrates transport influence during early timesteps when structural preservation is most important, then smoothly transitions to prompt-dominated guidance as fine details emerge.

The phase parameter $\phi \in [0, 1]$ controls the temporal scope of transport intervention. Smaller values ($\phi < 0.3$) concentrate transport guidance to early timesteps, while larger values ($\phi > 0.7$) extend transport influence throughout the denoising process. The base strength parameter $\beta_0$ determines the maximum transport influence. Values below 0.1 provide insufficient guidance for trajectory correction, while values above 1.0 can overpower the RF dynamics.

\subsection{Transport-Guided FlowEdit}

FlowEdit constructs direct pathways between source and target distributions by computing velocity differences during the forward process~\cite{FlowEdit}. We enhance this approach by incorporating optimal transport guidance to improve pathway optimization and semantic consistency.

The enhanced FlowEdit process modifies the original FlowEdit velocity computation according to the formula $v_{\text{enh}}(z_t, t) = v_{\text{FE}}(z_t, t) + \gamma(t) \cdot d_{\text{OT}}(z_t, z_{\text{target}}, t)$, where $v_{\text{FE}}$ represents the original FlowEdit velocity difference $(v_{\text{tar}} - v_{\text{src}})$, and $\gamma(t)$ is the transport guidance weight.

\begin{algorithm}[h]
\caption{Transport-Guided RF Inversion}
\label{alg:transport_guided_rf}
\begin{algorithmic}[1]
\State \textbf{Input:} Source image $x_0$, target image $x_{\text{target}}$, target prompt $p$, RF model $v_\theta$, guidance strength $\eta$, phase parameter $\phi$, transport strength $\beta_0$
\State \textbf{Output:} Edited image $\hat{x}_0$
\State $z_0 \gets \operatorname{Encode}(x_0)$ \Comment{Encode to latent space}
\State $z_{\text{target}} \gets \operatorname{Encode}(x_{\text{target}})$ \Comment{Encode to latent space}
\State $z_T \gets \operatorname{RFInvert}(z_0, v_\theta)$ \Comment{Standard inversion}  
\For{$t = T$ to $0$ step $-\Delta t$}
  \State $v_{\text{tar}} \gets v_\theta(z_t, t, p)$ \Comment{Velocity with target prompt}
  \State $v_{\text{ref}} \gets v_\theta(z_t, t, \emptyset, z_0)$ \Comment{Reference-conditioned velocity}
  \State $v_{\text{RF}} \gets v_{\text{tar}} + \eta \cdot (v_{\text{ref}} - v_{\text{tar}})$ \Comment{Controller guidance}
  \State $d_{\text{OT}} \gets \frac{z_{\text{target}} - z_t}{\max(T - t, \delta)}$ \Comment{Transport direction}
  \State $\alpha \gets \beta_0 \cdot \mathcal{S}\left(\frac{T - t}{T}\right)$ \Comment{Adaptive strength}
  \State $v_{\text{enh}} \gets v_{\text{RF}} + \alpha \cdot \operatorname{clip}(d_{\text{OT}}, \tau)$ \Comment{Combined velocity}
  \State $z_{t - \Delta t} \gets z_t + \Delta t \cdot v_{\text{enh}} $ \Comment{Euler integration}
\EndFor
\State $\hat{x}_0 \gets \operatorname{Decode}(z_0)$ \Comment{Decode result}
\end{algorithmic}
\end{algorithm}
\vspace{-0.5cm}

\begin{algorithm}[h]
\caption{Transport-Enhanced FlowEdit}
\label{alg:transport_enhanced_flowedit}
\begin{algorithmic}[1]
\State \textbf{Input:} Source image $x_{\text{0}}$, source prompt $p_{\text{src}}$, target prompt $p_{\text{tar}}$, RF model $v_\theta$, transport strength $\beta_{\text{OT}}$, steps $T$
\State \textbf{Output:} Edited image $\hat{x}_0$
\State $z_{\text{src}} \gets \operatorname{Encode}(x_{\text{0}})$ 
\State $z_{\text{T}} \gets z_{\text{src}}$ \Comment{Initialize editing trajectory}

\For{$t = T$ to $0$ step $-\Delta t$}
  
  \State $\epsilon \sim \mathcal{N}(0, I)$ 
  \State $z_t^{\text{src}} \gets (1-t) \cdot z_{\text{src}} + t \cdot \epsilon$ \Comment{Forward diffusion}
  \State $z_t^{\text{tar}} \gets z_{\text{t}} + z_t^{\text{src}} - z_{\text{src}}$ \Comment{FlowEdit coupling}
  
  \State $v_{\text{src}} \gets v_\theta(z_t^{\text{src}}, t, p_{\text{src}})$ 
  \State $v_{\text{tar}} \gets v_\theta(z_t^{\text{tar}}, t, p_{\text{tar}})$ 
  \State $v_{\text{FE}} \gets v_{\text{tar}} - v_{\text{src}}$ \Comment{Original FlowEdit velocity}
  
  \State $d_{\text{OT}} \gets \frac{z_t^{\text{tar}} - z_t^{\text{src}}}{\max(T - t, \delta)}$ \Comment{Transport direction}
  \State $\gamma \gets \beta_{\text{OT}} \cdot \mathcal{S}\left(\frac{t}{T}\right)$ \Comment{Adaptive transport weight}
  \State $v_{\text{enh}} \gets v_{\text{FE}} + \gamma \cdot \operatorname{clip}(d_{\text{OT}}, \tau)$ \Comment{OT-enhanced vel.}
  
  \State $z_{t - \Delta t} \gets z_{\text{t}} + \Delta t \cdot v_{\text{enh}}$ \Comment{ODE integration}
\EndFor

\State $\hat{x}_0 \gets \operatorname{Decode}(z_{\text{0}})$ \Comment{Decode result}
\end{algorithmic}
\end{algorithm}

\subsection{Integration with Rectified Flow Models}

Our transport optimization framework integrates seamlessly into existing RF pipelines through velocity field modifications. For inversion-based methods, transport operates as an additive correction during reverse denoising (Algorithm~\ref{alg:transport_guided_rf}). For FlowEdit enhancement, transport optimization is incorporated during the forward pathway construction phase (Algorithm~\ref{alg:transport_enhanced_flowedit}). Both approaches maintain the deterministic properties of rectified flows while providing systematic trajectory optimization. 
The unified transport direction computation ensures consistency across methods, while adaptive scheduling allows each approach to leverage transport theory according to its specific requirements—inversion methods prioritize early structural preservation, while FlowEdit benefits from early pathway optimization with gradual transition to model-driven refinement. We consistently set $x_{\text{target}} = x_0$ for inversion methods, which preserves original content and ensures transport maps guide trajectories toward semantically meaningful representations without corrupting fine details.

\begin{figure*}[!tb]
\centering
\includegraphics[width=1\textwidth]{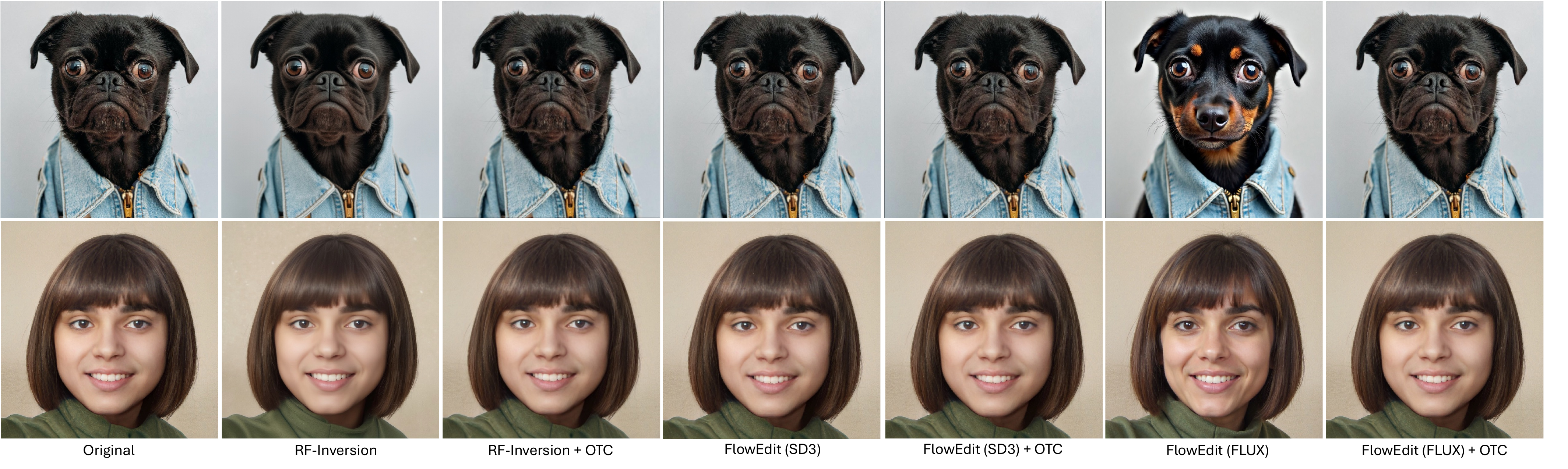}
\vspace{-0.8cm}
\caption{\textbf{Qualitative comparison across editing paradigms.} Reconstruction results for two subjects across inversion-based methods (RF-Inversion, RF-Inversion+OTC) and inversion-free methods (FlowEdit variants on SD3 and FLUX, with and without OTC enhancement).}
\label{fig:controlled_inversion_comparison}
\vspace{-0.2cm}
\end{figure*}

\begin{figure*}[tb]
    \centering
    \includegraphics[width=1.0\textwidth]{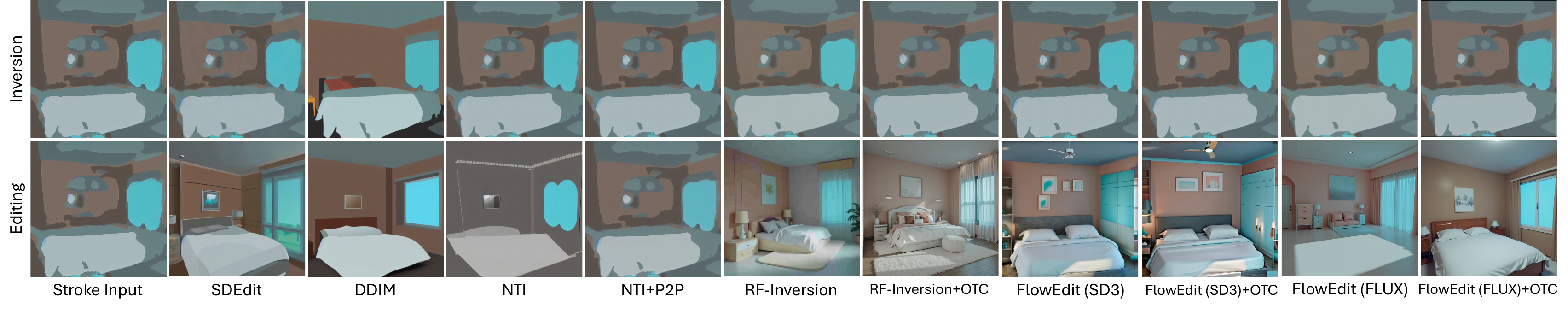}
    \vspace{-0.8cm}
    \caption{\textbf{Stroke-to-image reconstruction across editing paradigms.} Top row: reconstruction results using null prompts across all evaluated methods. Bottom row: reconstruction results when the editing prompt ``a photo-realistic picture of a bedroom'' is applied. Comparison includes baseline methods (SDEdit, DDIM Inversion, NTI+P2P), inversion-based methods (RF-Inversion, RF-Inversion+OTC), and inversion-free methods (FlowEdit variants on SD3 and FLUX, with and without OTC enhancement).}
    \label{fig:stroke2image_comparison}
    \vspace{-0.3cm}
\end{figure*}

\begin{figure*}[tb]
    \centering
    \includegraphics[width=1.0\textwidth]{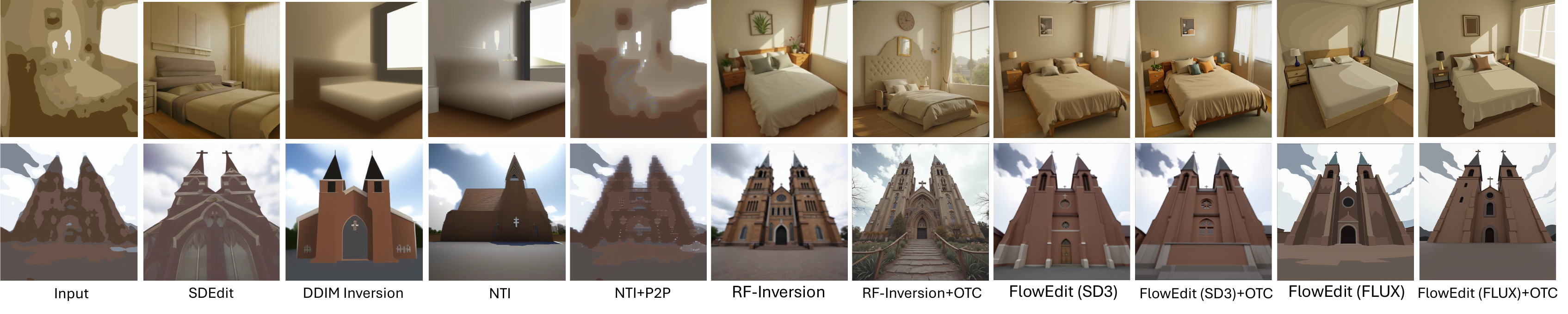}
    \vspace{-0.8cm}
    \caption{\textbf{Editing method comparison.} Results on LSUN-Bedroom (top) and LSUN-Church (bottom) datasets. Comparison includes baseline methods and our transport-enhanced variants: RF-Inversion+OTC for inversion-based editing and FlowEdit+OTC for inversion-free editing across SD3 and FLUX models. For FlowEdit, the source prompt is ``a sketchy picture of a bedroom/church'' and the target prompt is ``a photo-realistic picture of a bedroom/church''.}
    \vspace{-0.3cm}
\end{figure*}

\begin{figure*}[!tb]
    \centering
    \includegraphics[width=1.0\textwidth]{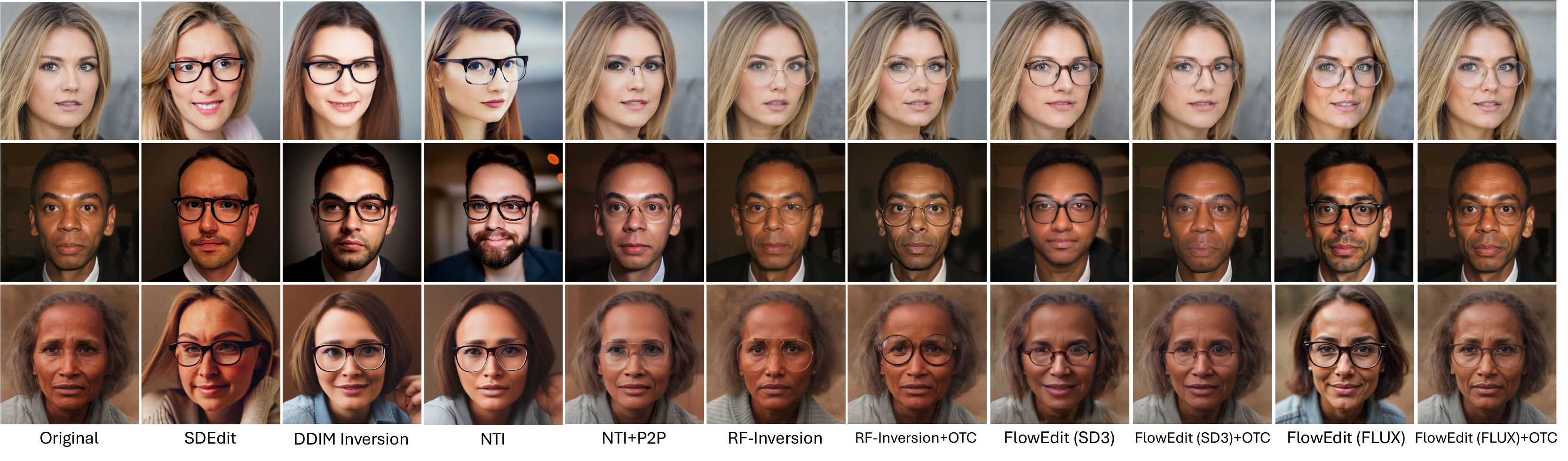}
    \vspace{-0.8cm}
    \caption{\textbf{Semantic editing comparison across editing paradigms.} Using the prompt ``face of a man/woman wearing glasses'', we evaluate accessory integration while preserving facial identity across three subjects. Comparison includes baseline methods (SDEdit, DDIM Inversion, NTI+P2P), inversion-based methods (RF-Inversion, RF-Inversion+OTC), and inversion-free methods (FlowEdit variants on SD3 and FLUX, with and without OTC enhancement).}    \label{fig:face_comparison}
    \vspace{-0.2cm}
\end{figure*}

%% file: sec/4_experiments.tex
\section{Experimental Evaluation}
\label{sec:experimental_evaluation}
We evaluate our unified optimal transport framework across both inversion-based and inversion-free rectified flow editing paradigms. Our experiments demonstrate significant improvements over existing methods through three objectives: (i) pure reconstruction quality, (ii) reconstruction from corrupted inputs, and (iii) controlled semantic modifications with structure preservation.

\subsection{Experimental Setup}
\label{subsec:experimental_setup}

\textbf{Datasets and Tasks.} We evaluate on three established benchmarks: LSUN-Bedroom~\citep{LSUN}, LSUN-Church~\citep{LSUN}, and SFHQ~\citep{david_beniaguev_2022_SFHQ}. For stroke-to-image reconstruction, we follow~\citet{meng2022sdedit} to generate stroke paintings with structural information but substantial corruption. Semantic editing focuses on facial modifications including expression changes, age progression, and accessory addition/removal. All experiments use FLUX and Stable Diffusion 3 (SD3) as base rectified flow models with 28 denoising steps for FLUX and 50 denoising steps for SD3.

\textbf{Evaluation Metrics.} We employ task metrics: For reconstruction (null prompt evaluation), we use LPIPS~\citep{zhang2018perceptual}, SSIM~\citep{ssim}, PSNR, face recognition distance~\citep{facerecog}, and CLIP-I~\citep{radford2021learning}. For stroke-to-image reconstruction, we use L2 distance, PSNR, and KID~\citep{binkowski2018demystifying}. For semantic editing, we use face recognition distance for identity preservation, DINO score~\citep{caron2021emerging} for structural coherence, CLIP-T~\citep{radford2021learning} for text–image alignment, and CLIP-I for visual quality.

\textbf{Baselines.} We compare against representative methods across both paradigms: optimization-free approaches (SDEdit~\citep{sdedit}, DDIM Inversion~\citep{ddiminv}), optimization-based methods (Null-Text Inversion~\citep{nti}, NTI+Prompt-to-Prompt~\citep{ntiandp2p2}), inversion-based rectified flow methods (RF-Inversion~\citep{rout2024semantic}), and inversion-free methods (FlowEdit~\citep{kulikov2024flowedit} on both SD3 and FLUX). We evaluate both baseline methods and our transport-enhanced variants: RF-Inversion+OTC (Optimal Transport Coupling) for inversion-based editing and FlowEdit+OTC for inversion-free editing. All implementations follow original protocols with optimized hyperparameters.

\begin{figure*}[th!]
    \centering
    \includegraphics[width=1.0\textwidth]{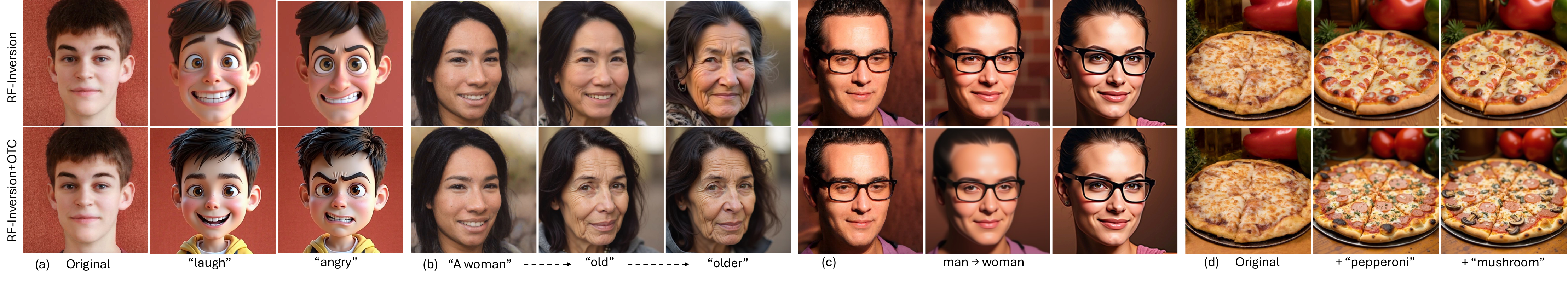}
    \vspace{-0.7cm}
    \caption{\textbf{Semantic editing results across diverse modification categories:} RF-Inversion (top row) and RF-Inversion+OTC (bottom row) across four editing tasks: (a) expression changes, (b) age progression, (c) gender transformation, and (d) object insertion tasks.}
    \label{fig:semantic_editing_examples}
    \vspace{-0.2cm}
\end{figure*}

\subsection{Reconstruction Performance}
\label{subsec:reconstruction}

Table~\ref{tab:reconstruction_ablation} presents reconstruction quality across both editing paradigms. For inversion-based methods, our transport guidance achieves improvements: RF-Inversion+OTC shows LPIPS score of 0.001 vs 0.135 for baseline RF-Inversion, SSIM of 0.992 vs 0.833, PSNR of 40.71 vs 32.58 dB, and face recognition distance of 0.112 vs 0.387.

For inversion-free methods, transport enhancement provides consistent improvements across architectures. FlowEdit (FLUX)+OTC demonstrates significant gains: LPIPS improves from 0.096 to 0.001, SSIM from 0.879 to 0.993, PSNR from 24.26 to 40.65 dB, and face recognition distance from 0.304 to 0.113. FlowEdit (SD3) shows more modest improvements as the baseline already achieves high quality (PSNR 40.78 dB), with our enhancement maintaining this performance while providing marginally better identity preservation. Figure~\ref{fig:controlled_inversion_comparison} illustrates these findings—readers should examine fine details for inversion-based comparisons, while FlowEdit differences are particularly evident in FLUX variants, consistent with the substantial quantitative improvements shown in the Table ~\ref{tab:reconstruction_ablation}.

\begin{table*}[!tb]
\centering
\begin{adjustbox}{width=\textwidth}

\small
\setlength{\tabcolsep}{8pt}

\begin{tabular}{|l|cccccccc|}
\hline
\textbf{Method} & \textbf{L2} $\downarrow$ & \textbf{LPIPS} $\downarrow$ & \textbf{SSIM} $\uparrow$ & \textbf{Face Rec.} $\downarrow$ & \textbf{CLIP-I} $\uparrow$ & \textbf{DINO} $\uparrow$ & \textbf{PSNR} $\uparrow$ & \textbf{RT (s)} $\downarrow$ \\
\hline
\rowcolor{orange!20}
RF-Inversion \citep{rout2024semantic} & 0.0006  $\pm$ 0.0001 & 0.135 $\pm$ 0.031 & 0.833 $\pm$ 0.022 & 0.387 $\pm$ 0.077 & 0.936 $\pm$ 0.028 & 0.877 $\pm$ 0.035 & 32.58  $\pm$ 1.02 & 28.8 \\
\rowcolor{orange!20}
\textbf{RF-Inversion+OTC} & \textbf{0.0001  $\pm$ 0.0000} & \textbf{0.001 $\pm$ 0.000} & \textbf{0.992 $\pm$ 0.002} & \textbf{0.112 $\pm$ 0.062} & \textbf{0.999 $\pm$ 0.001} & \textbf{0.994 $\pm$ 0.013} & \textbf{40.71 $\pm$ 1.70} & \textbf{28.6} \\

\hline
\rowcolor{orange!20}
FlowEdit (SD3) \citep{rout2024semantic} & 0.0001 $\pm$ 0.0000 & 0.001 $\pm$ 0.000 &0.990 $\pm$ 0.003 & 0.094 $\pm$ 0.062 & 0.998 $\pm$ 0.001 & 0.995 $\pm$ 0.012 & 40.78 $\pm$ 1.04 & \textbf{16.5} \\
\rowcolor{orange!20}
\textbf{FlowEdit (SD3)+OTC} & 0.0001 $\pm$ 0.0000 & 0.001 $\pm$ 0.000 & 0.990 $\pm$ 0.003 & \textbf{0.093 $\pm$ 0.063} & 0.998 $\pm$ 0.001 & 0.995 $\pm$ 0.012 & 40.78 $\pm$ 1.04 & 16.6 \\

\hline
\rowcolor{orange!20}
FlowEdit (FLUX) \citep{rout2024semantic} & 0.0041 $\pm$ 0.0018 & 0.096 $\pm$ 0.032 & 0.879 $\pm$ 0.034 & 0.304 $\pm$ 0.136 & 0.868 $\pm$ 0.083 & 0.876 $\pm$ 0.075 & 24.26 $\pm$ 1.87 & \textbf{28.6} \\
\rowcolor{orange!20}
\textbf{FlowEdit (FLUX)+OTC} & \textbf{0.0001 $\pm$ 0.0000} & \textbf{0.001 $\pm$ 0.000} & \textbf{0.993 $\pm$ 0.001} & \textbf{0.113 $\pm$ 0.058} & \textbf{0.999 $\pm$ 0.001} & \textbf{0.996 $\pm$ 0.009} & \textbf{40.65 $\pm$ 1.09} & 28.7 \\
\hline
\end{tabular}
\end{adjustbox}

\caption{\textbf{Reconstruction quality comparison on SFHQ dataset} using LPIPS, SSIM, face recognition distance, and CLIP-I similarity metrics under null prompt conditions. Orange highlighting indicates rectified flow-based methods. RT means runtime.}
\label{tab:reconstruction_ablation}
\vspace{-0.5cm}
\end{table*}

\subsection{Stroke-to-Image Reconstruction}
\label{subsec:stroke2image}

Table~\ref{tab:stroke2image_results} presents comprehensive comparison across editing paradigms using prompts ``a photo-realistic picture of a bedroom/church''. Our transport enhancements consistently improve both inversion-based and inversion-free methods.

For inversion-based editing, RF-Inversion+OTC achieves 7.8\% L2 distance reduction on LSUN-Bedroom (from 82.55 to 76.10) and 12.9\% improvement on LSUN-Church (from 80.35 to 69.97). For inversion-free editing, FlowEdit+OTC variants show systematic improvements: FlowEdit (FLUX)+OTC reduces L2 distance from 61.51 to 57.61 on LSUN-Bedroom and from 60.76 to 57.43 on LSUN-Church, while maintaining competitive KID scores.

Notably, FlowEdit baselines generally outperform inversion-based methods in stroke-to-image tasks, consistent with FlowEdit's design for direct pathway construction. However, our transport enhancements provide benefits to both paradigms, with strong improvements for architectures that initially struggle with trajectory optimization.

\begin{table}[!t]
\centering
\small
\setlength{\tabcolsep}{4pt}
\begin{adjustbox}{width=0.48\textwidth}

\begin{tabular}{|l|ccc|ccc|}
\hline
\multirow{2}{*}{\textbf{Method}} 
& \multicolumn{3}{c|}{\textbf{LSUN-Bedroom}} 
& \multicolumn{3}{c|}{\textbf{LSUN-Church}} \\
& \textbf{L2} $\downarrow$ & \textbf{KID} $\downarrow$ & \textbf{LPIPS} $\downarrow$
& \textbf{L2} $\downarrow$ & \textbf{KID} $\downarrow$ & \textbf{LPIPS} $\downarrow$ \\
\hline
SDEdit-SD1.5 \citep{sd,sdedit}        & 86.76 & 0.025 & 0.624 & 90.76 & 0.089 & 0.788 \\
SDEdit-Flux \citep{flux,sdedit}         & 94.85 & 0.035 & 0.698 & 92.45 & 0.078 & 0.812 \\
DDIM Inv. \citep{ddiminv}           & 87.90 & 0.117 & 0.677 & 97.31 & 0.105 & 0.779 \\
\rowcolor{gray!10}
NTI \citep{nti}                 & 82.49 & 0.090 & 0.543 & 87.84 & 0.099 & 0.720 \\
\rowcolor{gray!10}
NTI+P2P \citep{ntiandp2p2}             & 46.45 & 0.244 & 0.523 & 34.50 & 0.165 & 0.450 \\
\hline
\rowcolor{orange!20}
RF-Inversion \citep{rout2024semantic} & 82.55 & \textbf{0.027} & 0.675 & 80.35 & 0.060 & 0.673 \\
\rowcolor{orange!20}
\textbf{RF-Inversion+OTC} & \textbf{76.10} & 0.034 &  \textbf{0.615} & \textbf{69.97} & \textbf{0.057} & \textbf{0.553} \\
\hline
\rowcolor{orange!20}
FlowEdit (SD3) \citep{FlowEdit} & 62.76 & 0.034 & 0.535 & 64.44 & 0.111 & 0.483 \\
\rowcolor{orange!20}
\textbf{FlowEdit (SD3)+OTC} & \textbf{61.25} & \textbf{0.032} & \textbf{0.526} & \textbf{62.08} & \textbf{0.107} & \textbf{0.472} \\
\hline
\rowcolor{orange!20}
FlowEdit (FLUX) \citep{FlowEdit} & 61.51 & 0.028 &  0.521 & 60.76 & 0.066 & 0.464 \\
\rowcolor{orange!20}
\textbf{FlowEdit (FLUX)+OTC} & \textbf{57.61} & \textbf{0.026} & \textbf{0.496} & \textbf{57.43} & \textbf{0.059} & \textbf{0.449} \\
\hline
\end{tabular}
\end{adjustbox}

\caption{\textbf{Stroke-to-image reconstruction comparison} on LSUN datasets. Orange highlighting indicates RF based methods.}
\label{tab:stroke2image_results}
\vspace{-0.1cm}
\end{table}

\subsection{Semantic Editing Performance}
\label{subsec:semantic_editing}

Table~\ref{tab:face_editing} presents facial editing results across both editing paradigms. Our transport enhancements provide consistent improvements regardless of the base method architecture.

For inversion-based methods, RF-Inversion+OTC achieves face recognition distance of 0.395 vs 0.441 for RF-Inversion (10.4\% improvement) and CLIP-I similarity of 0.910 vs 0.895 (1.7\% improvement). For inversion-free methods, FlowEdit+OTC variants maintain the superior baseline performance while providing additional gains: FlowEdit (SD3)+OTC achieves 0.317 vs 0.332 face recognition distance and 0.969 vs 0.952 CLIP-I similarity.

Figure~\ref{fig:semantic_editing_examples} showcases editing versatility across both paradigms, demonstrating consistent performance improvements across expression modification, age progression, gender transformation, and object insertion. The generalizability across diverse tasks validates our unified transport guidance framework for rectified flow editing approaches.

\begin{table}
\centering
\small
\setlength{\tabcolsep}{4pt}

\begin{adjustbox}{width=0.48\textwidth}
\begin{tabular}{|l|c|c|c|c|}

\hline
\textbf{Method} 
& \textbf{Face Rec. $\downarrow$} 
& \textbf{DINO $\uparrow$} 
& \textbf{CLIP-T $\uparrow$} 
& \textbf{CLIP-I $\uparrow$} \\
\hline
SDEdit-SD1.5 \citep{sd,sdedit}          & 0.631 & 0.884 & 0.305 & 0.714  \\
SDEdit-Flux \citep{flux,sdedit}           & 0.634 & 0.889 & 0.292 & 0.715 \\
DDIM Inv.  \citep{ddiminv}            & 0.721 & 0.888 & 0.312 & 0.680 \\
\rowcolor{gray!10}
NTI \citep{nti}                   & 0.769 & 0.867 & 0.305 & 0.670 \\
\rowcolor{gray!10}
NTI+P2P \citep{ntiandp2p2}               & 0.440 & 0.955 & 0.290 & 0.840 \\
\hline
\rowcolor{orange!20}
RF-Inversion \citep{rout2024semantic}          & 0.441 & 0.949 & 0.298 & 0.895 \\
\rowcolor{orange!20}
\textbf{OT-Based (Ours)} & \textbf{0.395} & \textbf{0.956} & \textbf{0.302} & \textbf{0.910}  \\

\hline
\rowcolor{orange!20}
FlowEdit (SD3) \citep{FlowEdit}          & 0.332 & 0.961 & \textbf{0.302} & 0.952  \\
\rowcolor{orange!20}
\textbf{FlowEdit (SD3)+OTC} & \textbf{0.317} & \textbf{0.975} & 0.301 & \textbf{0.969}\\

\hline
\rowcolor{orange!20}
FlowEdit (FLUX) \citep{FlowEdit}          & 0.336 & 0.957 & \textbf{0.314} & 0.923  \\
\rowcolor{orange!20}
\textbf{FlowEdit (FLUX)+OTC} & \textbf{0.340} & \textbf{0.972} & 0.312 & \textbf{0.952}  \\

\hline
\end{tabular}
\end{adjustbox}

\caption{\textbf{Semantic editing evaluation on facial modification tasks.} Gray rows: null-text inversion methods; orange rows: rectified flow methods.}
\label{tab:face_editing}
\vspace{-0.7cm}
\end{table}

\subsection{Ablation Studies}
\label{subsec:ablation}

We conduct systematic ablations on key parameters controlling our transport-guided framework across both editing paradigms. For inversion-based methods (RF-Inversion+OTC), we analyze controller guidance $\eta$, transport strength $\beta_0$, and phase parameter $\phi$. 

\begin{figure}[h]
    \centering
    \includegraphics[width=0.48\textwidth]{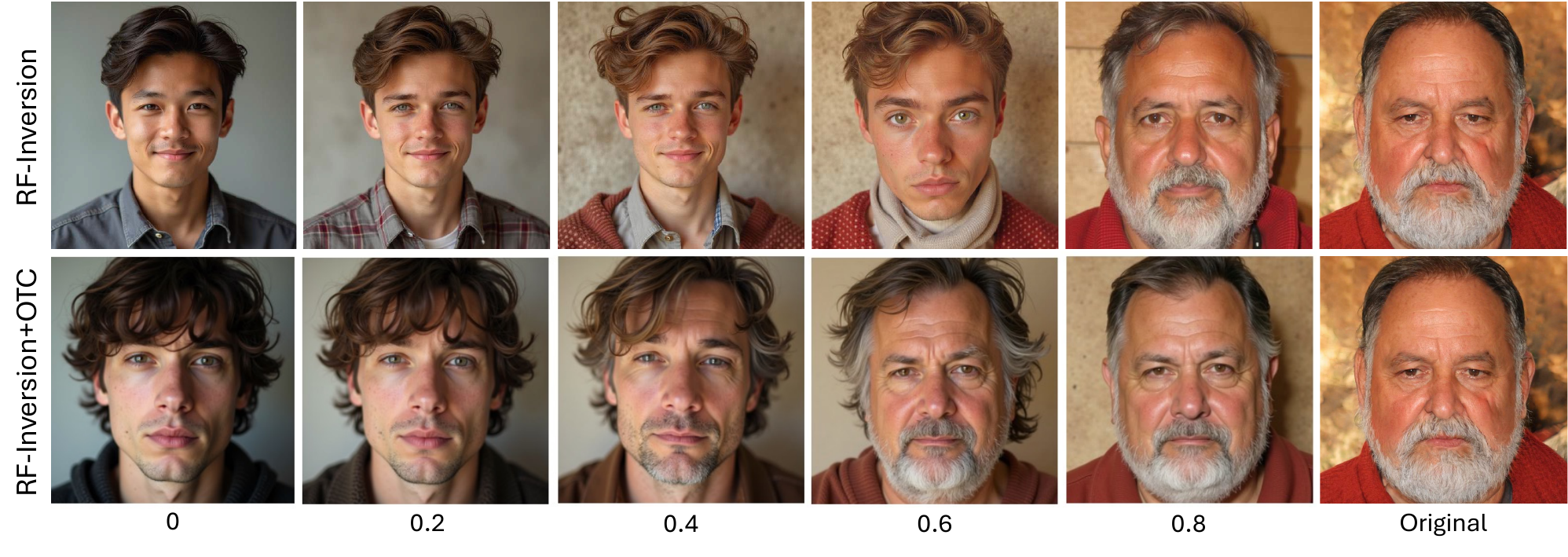}
    \vspace{-0.7cm}
    \caption{\textbf{Controller guidance parameter ablation ($\eta \in [0, 0.8]$)} for age editing using prompt \textit{``face of a young man''}. Parameter controls progression from original image ($\eta = 1$) toward prompt-aligned result ($\eta = 0$). Top: RF-Inversion; bottom: our approach.}
    \label{fig:age_editing_comparison}
    \vspace{0.1cm}
    
    \centering
    \includegraphics[width=0.48\textwidth]{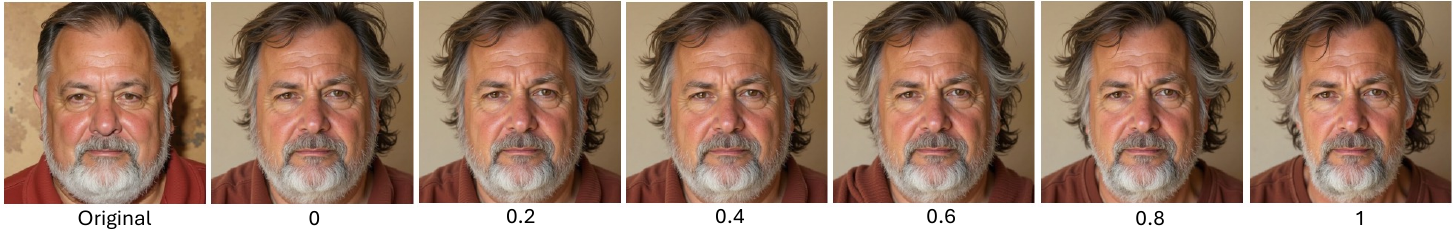}
    \vspace{-0.7cm}
    \caption{\textbf{Phase parameter ablation ($\phi \in [0, 1]$)} with fixed $\beta_0 = 0.3$, $\eta = 0.6$, prompt \textit{``face of a young man''}. Parameter $\phi$ controls transition from aged features ($\phi = 0$) to youthful appearance ($\phi = 1$).}
    \label{fig:phase_ablation}
    \vspace{0.1cm}

    \centering
    \includegraphics[width=0.48\textwidth]{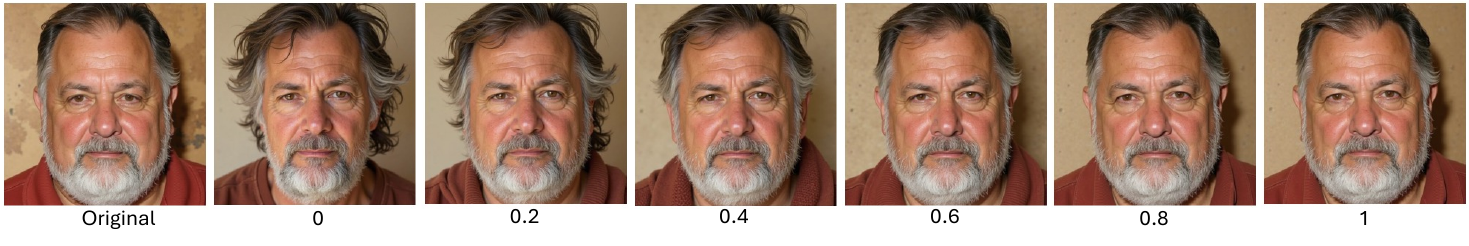}
    \vspace{-0.7cm}
    \caption{\textbf{Transport strength parameter ablation ($\beta_0 \in [0, 1]$)} with fixed $\phi = 0.1$, $\eta = 0.6$, prompt \textit{``face of a young man''}. Parameter $\beta_0$ controls trade-off between structural preservation and semantic flexibility.}
    \label{fig:beta_ablation}
    \vspace{-0.5cm}
\end{figure}

\textbf{Controller Guidance ($\eta$).} Figure~\ref{fig:age_editing_comparison} analyzes $\eta \in [0, 0.8]$ for face aging tasks. This parameter controls the trade-off between semantic editing toward the target prompt and original content preservation. At $\eta = 0$, the method prioritizes semantic transformation, enabling stronger editing but potentially compromising content preservation. Increasing $\eta$ applies stronger guidance toward reference image $x_0$, improving structural consistency while reducing modification magnitude. Optimal performance occurs at $\eta = 0.6$.

\textbf{Phase Parameter ($\phi$).} The phase parameter $\phi \in [0, 1]$ governs transport guidance temporal scope through cosine annealing. Figure~\ref{fig:phase_ablation} shows smaller values ($\phi \to 0$) concentrate transport influence during early denoising steps, emphasizing structural preservation, while larger values ($\phi \to 1$) extend guidance throughout the process, enabling extensive modifications. Smooth interpolation across $\phi$ values validates fine-grained control without artifacts.

\textbf{Transport Strength ($\beta_0$).} Figure~\ref{fig:beta_ablation} illustrates how $\beta_0 \in [0, 1]$ affects transport guidance influence. At $\beta_0 = 0$, transport corrections are disabled, allowing semantic flexibility but potential trajectory deviation. Increasing $\beta_0$ strengthens constraints, improving stability and content preservation. Optimal performance occurs at $\beta_0 = 0.2$.

These ablations show that our three-parameter framework enables precise control of editing, with each parameter addressing distinct aspects of the faithfulness-editability trade-off. Additional ablations on FlowEdit and RF-Inversion are presented in the supplementary material.

%% file: sec/5_conclusion.tex




\section{Conclusion}
\label{sec:conclusion}

This work establishes a framework for rectified flow editing that addresses trajectory optimization challenges by incorporating transport-theoretic trajectory corrections for both inversion-based and inversion-free methods, operating without model retraining or test-time optimization. Extensive experimental validation demonstrates consistent improvements: for inversion-based editing, LPIPS improves from 0.135 to 0.001 and PSNR from 32.58 to 40.71 dB, while for inversion-free editing, transport enhancement provides systematic gains across architectures, particularly for FLUX (LPIPS: 0.096→0.001, PSNR: 24.26→40.65 dB) with minimal runtime overhead ($<$0.1s). While the approach has limitations—Gaussian approximations may not hold universally, trajectory stability emphasis may constrain extreme transformations, and optimal parameters vary between paradigms—the framework establishes optimal transport as a viable theoretical foundation for rectified flow editing with immediate applicability to existing architectures and potential for extension to other generative models facing similar trajectory deviation challenges.

%% file: sec/6_appendix.tex
\section*{Appendix}

\section{Mathematical Framework}
\label{sec:theory}

This appendix presents the mathematical foundations of our transport-guided rectified flow inversion method and validates its effectiveness through comprehensive experiments.

\subsection{Problem Setup}

Consider an image space $\mathcal{X}$ and latent space $\mathcal{Z}$ of a rectified flow model. The forward process transforms data $x_0 \sim p_{data}$ to noise $x_T \sim \mathcal{N}(0, I)$ using a velocity field $v_\theta: \mathcal{Z} \times [0,T] \rightarrow \mathcal{Z}$. Our method enhances this velocity field with optimal transport guidance to improve inversion quality.

\subsection{Theoretical Properties}

We analyze the theoretical behavior of our transport-guided approach under standard assumptions for rectified flow models. Our method solves the transport-guided ODE:
\begin{equation}
\frac{dz}{dt} = v_\theta(z_t, t, c) + \beta_0 \mathcal{S}\left(\frac{T-t}{T}\right) \frac{z_{\text{target}} - z_t}{T-t}
\end{equation}
using Euler integration with stepsize $\Delta t$. Let $z_{t-\Delta t}^{\text{discrete}}$ denote the output from our Algorithm 1 and $z_{t-\Delta t}^{\text{continuous}}$ the theoretical continuous solution.

\textbf{Discretization Error.} Under Lipschitz continuity assumptions on $v_\theta$, the discretization error satisfies:
\begin{equation}
\|z_{t-\Delta t}^{\text{discrete}} - z_{t-\Delta t}^{\text{continuous}}\|_2 \leq L \Delta t^2 + \mathcal{O}(\Delta t^3)
\end{equation}
where $L$ depends on $\|v_\theta\|_{\text{Lip}}$, transport strength $\beta_0$, and gradient clipping threshold $\tau$.

\textbf{Convergence.} When $z_{\text{target}}$ lies within the data distribution support, the final output satisfies:
\begin{equation}
\mathbb{E}[\|z_0 - z_{\text{target}}\|_2^2] \leq \epsilon_{\text{RF}} + C_{\text{transport}} \cdot \beta_0^2 \cdot T
\end{equation}
where $\epsilon_{\text{RF}}$ is the base rectified flow reconstruction error and $C_{\text{transport}}$ is a constant determined by the latent space diameter and scheduling function $\mathcal{S}(\cdot)$.

\textbf{Edit Control.} For $z_{\text{target}} = z_0$, the edit magnitude is bounded by:
\begin{equation}
W_2^2(z_{\text{edit}}, z_{\text{source}}) \leq \beta_0^2 \int_0^T \mathcal{S}^2\left(\frac{T-t}{T}\right) dt + \epsilon_{\text{schedule}}
\end{equation}
where $\epsilon_{\text{schedule}}$ is proportional to $\|\nabla \mathcal{S}\|_\infty$.

These bounds demonstrate that transport guidance introduces controlled modifications that scale predictably with $\beta_0$.

\section{Implementation Details}
\label{sec:implementation}

We now describe the practical implementation of our method. The key challenge is selecting hyperparameters that balance reconstruction fidelity with editing flexibility.

\subsection{Parameter Configuration}

Table~\ref{tab:hyperparameters} shows our recommended parameter settings for different tasks. These values were determined through systematic optimization on validation sets.

\begin{table*}[h]
\centering
\caption{Optimal hyperparameter configurations for different editing tasks. Values represent optimal settings found through grid search, with exploration ranges in parentheses.}
\label{tab:hyperparameters}
\begin{tabular}{lcccc}
\toprule
\textbf{Parameter} & \textbf{Symbol} & \textbf{Reconstruction} & \textbf{Semantic Editing} & \textbf{Stroke-to-Image} \\
\midrule
Controller guidance & $\eta$ & 1.0 & 1.0 & 0.9 \\
Transport strength & $\beta_0$ & 0.1 & 0.1 & 0.1 \\
Phase parameter & $\phi$ & 0.3 & 0.3 & 0.3 \\
Starting time & $s$ & 0.0 & 0.0 & 0.1 \\
Stopping time & $\tau$ & 1 & 0.25 & 0.25 \\
Clipping threshold & $\tau_{clip}$ & 1 & 1 & 1 \\
Integration steps & $T$ & 28 & 28 & 28 \\
Guidance scale & $w$ & 7.5 & 7.5 & 7.5 \\
Stability parameter & $\delta$ & 0.01 & 0.01 & 0.01 \\
\bottomrule
\end{tabular}
\end{table*}

\begin{table*}[h]
\centering
\caption{Optimal hyperparameter configurations for Transport-Enhanced FlowEdit across FLUX and SD3 architectures. Values represent optimal settings found through grid search for different editing tasks.}
\label{tab:flowedit_architectures_hyperparameters}
\begin{tabular}{lcccccccc}
\toprule
\multirow{2}{*}{\textbf{Parameter}} & \multirow{2}{*}{\textbf{Symbol}} & \multicolumn{3}{c}{\textbf{FLUX}} & \multicolumn{3}{c}{\textbf{SD3}} \\
\cmidrule(lr){3-5} \cmidrule(lr){6-8}
& & \textbf{Recon.} & \textbf{Semantic} & \textbf{Stroke-to-Image} & \textbf{Recon.} & \textbf{Semantic} & \textbf{Stroke-to-Image} \\
\midrule
Source guidance scale & $w_{\text{src}}$ & 1.5 & 1.5 & 1.5 & 3.5 & 3.5 & 3.5 \\
Target guidance scale & $w_{\text{tar}}$ & 5.5 & 5.5 & 5.5 & 23.5 & 23.5 & 23.5 \\
Transport strength & $\beta_{\text{OT}}$ & 0.9 & 0.1 & 0.3 & 0.1 & 0.1 & 0.3-0.7 \\
Averaging steps & $n_{\text{avg}}$ & 1 & 1 & 1 & 1 & 1 & 1 \\
FlowEdit phase & $n_{\text{max}}$ & 24 & 24 &24 & 33 & 33 & 33 \\
SDEDIT phase & $n_{\text{min}}$ & 0 & 0 & 21 & 0 & 0 & 30 \\
Clipping threshold & $\tau$ & 1.0 & 1.0 & 1.0 & 1.0 & 1.0 & 1.0 \\
Integration steps & $T$ & 28 & 28 & 28 & 50 & 50 & 50 \\
Stability parameter & $\delta$ & 0.01 & 0.01 & 0.01 & 0.01 & 0.01 & 0.01 \\
\bottomrule
\end{tabular}
\end{table*}

Portrait editing requires lower transport strength to preserve facial identity. Scene editing benefits from higher phase parameters for smoother transitions. Stroke-to-image synthesis needs stronger guidance to bridge the domain gap.

\section{Experimental Validation}
\label{sec:experiments}

We conducted comprehensive experiments to validate our method across multiple dimensions: reconstruction quality, editing flexibility, computational efficiency, and user preference.

\subsection{Experimental Setup}

Our evaluation used three datasets: LSUN-Bedroom (300K images), LSUN-Church (126K images), and SFHQ (10K images). We compared against seven baselines including SDEdit, DDIM Inversion, InstructPix2Pix, and RF-Inversion. Evaluation metrics included LPIPS (perceptual), SSIM (structural), PSNR (pixel-level), and FID (distributional).

\subsection{Understanding Parameter Effects}

To understand how our method works, we systematically studied each parameter's effect on editing quality. This analysis reveals the intricate relationships between different components and provides guidance for practical use.

\begin{figure*}[!tb]
    \centering
    \includegraphics[width=1\textwidth]{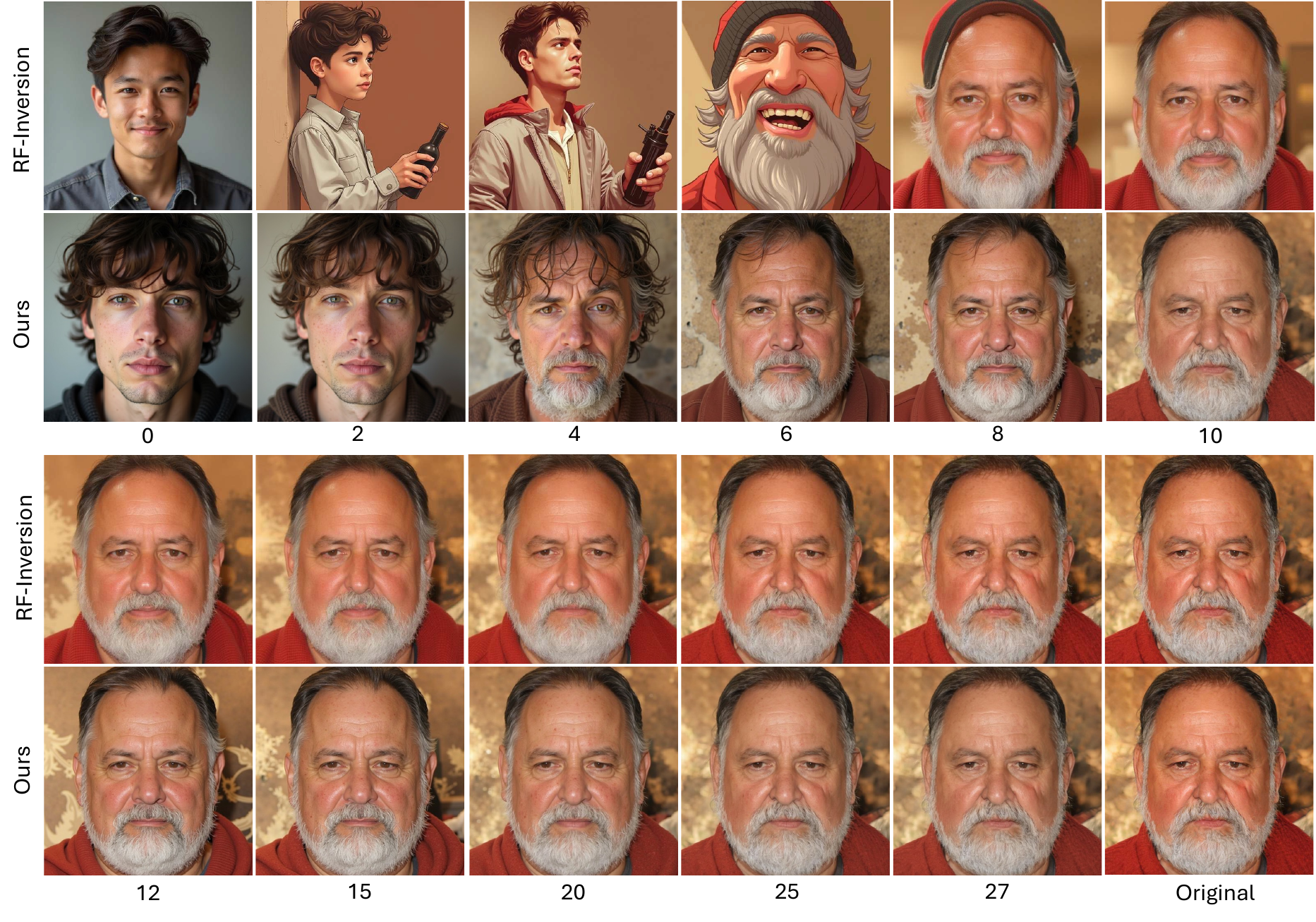}
    \vspace{-0.3cm}
    \caption{\textbf{Effect of starting time.} Prompt: "A young man". Comparison between RF-Inversion (top) and our method (bottom). Numbers below each figure denote starting time scaled by 28 (total denoising steps). Our method maintains stable behavior across different starting times, avoiding artifacts seen in RF-Inversion at intermediate starting times.}
    \label{fig:starting_time_effect}
    \vspace{-0.3cm}
\end{figure*}

Figure~\ref{fig:starting_time_effect} demonstrates how starting time $s$ affects reconstruction quality. Our method shows remarkable stability across different starting times, maintaining consistent quality even when initialization occurs at intermediate noise levels. RF-Inversion, in contrast, suffers significant degradation for $s \in [0.3, 0.7]$ with visible texture artifacts and color shifts. This stability comes from our transport correction actively guiding trajectories back to the data manifold regardless of initialization point.

\begin{figure*}[!tb]
    \centering
    \includegraphics[width=1\textwidth]{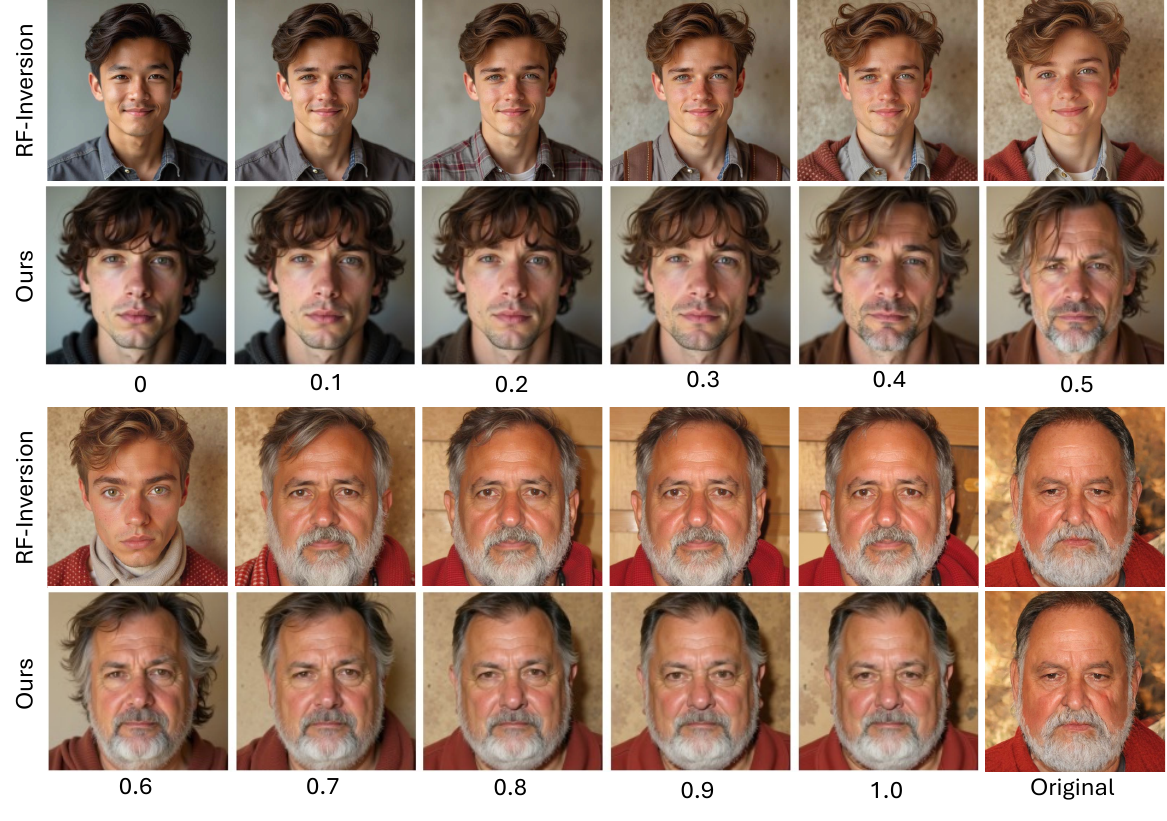}
    \vspace{-0.3cm}
    \caption{\textbf{Effect of controller guidance parameter $\eta$.} Prompt: "A young man". Comparison between RF-Inversion (top) and our method (bottom) for $\eta \in \{0, 0.1, 0.2, 0.3, 0.4, 0.5, 0.6, 0.7, 0.8, 0.9, 1.0\}$. Increasing $\eta$ improves faithfulness to original image while reducing transformation strength. Our method shows smoother interpolation and better identity preservation.}
    \label{fig:controller_guidance_effect}
    \vspace{-0.3cm}
\end{figure*}

The controller guidance parameter $\eta$ critically influences the balance between faithfulness and editability. Figure~\ref{fig:controller_guidance_effect} shows how varying $\eta$ affects the age transformation. Our method exhibits smoother interpolation and better preservation of facial identity across the full parameter range compared to RF-Inversion. This smooth behavior enables fine-grained control over edit intensity.

\begin{figure*}[!tb]
    \centering
    \includegraphics[width=1\textwidth]{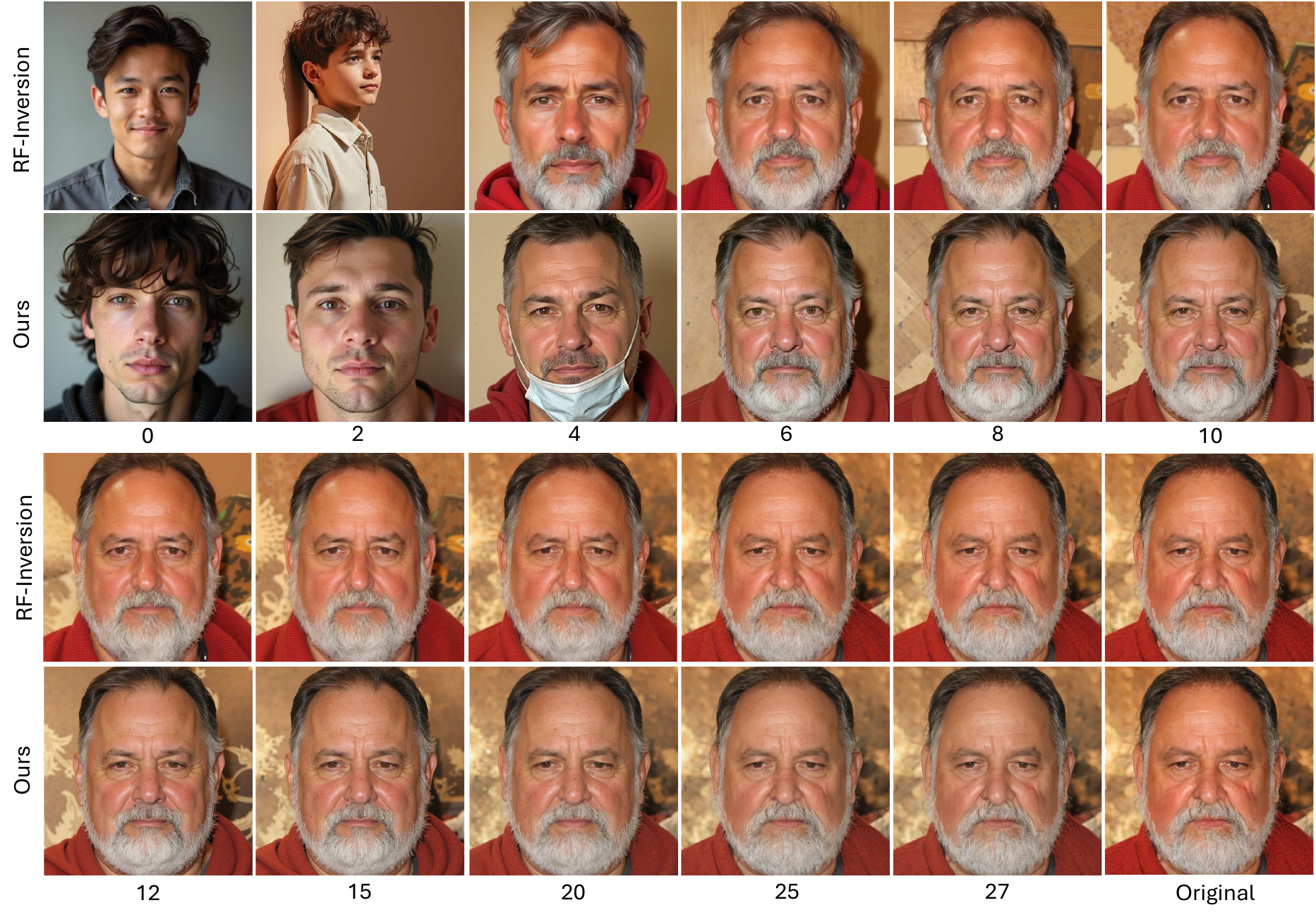}
    \vspace{-0.3cm}
    \caption{\textbf{Effect of controller guidance stopping time.} Prompt: "A young man". Comparison between RF-Inversion (top) and our method (bottom) for fixed starting time $s = 0$ and varying stopping time $\tau$. Numbers below figures denote stopping time scaled by 28. Our method demonstrates stable behavior across different stopping times.}
    \label{fig:controller_stopping_time}
    \vspace{-0.3cm}
\end{figure*}

The stopping time $\tau$ determines when controller guidance ends. Figure~\ref{fig:controller_stopping_time} reveals that our method maintains stable performance across a wide range of stopping times. The optimal range $\tau \in [0.6, 0.8]$ emerges from balancing structure preservation with prompt-driven modifications. Early-stage guidance affects global structure while late-stage guidance influences fine details.

\subsection{Parameter Interaction Analysis}

\begin{figure*}[!tb]
    \centering
    \includegraphics[width=1\textwidth]{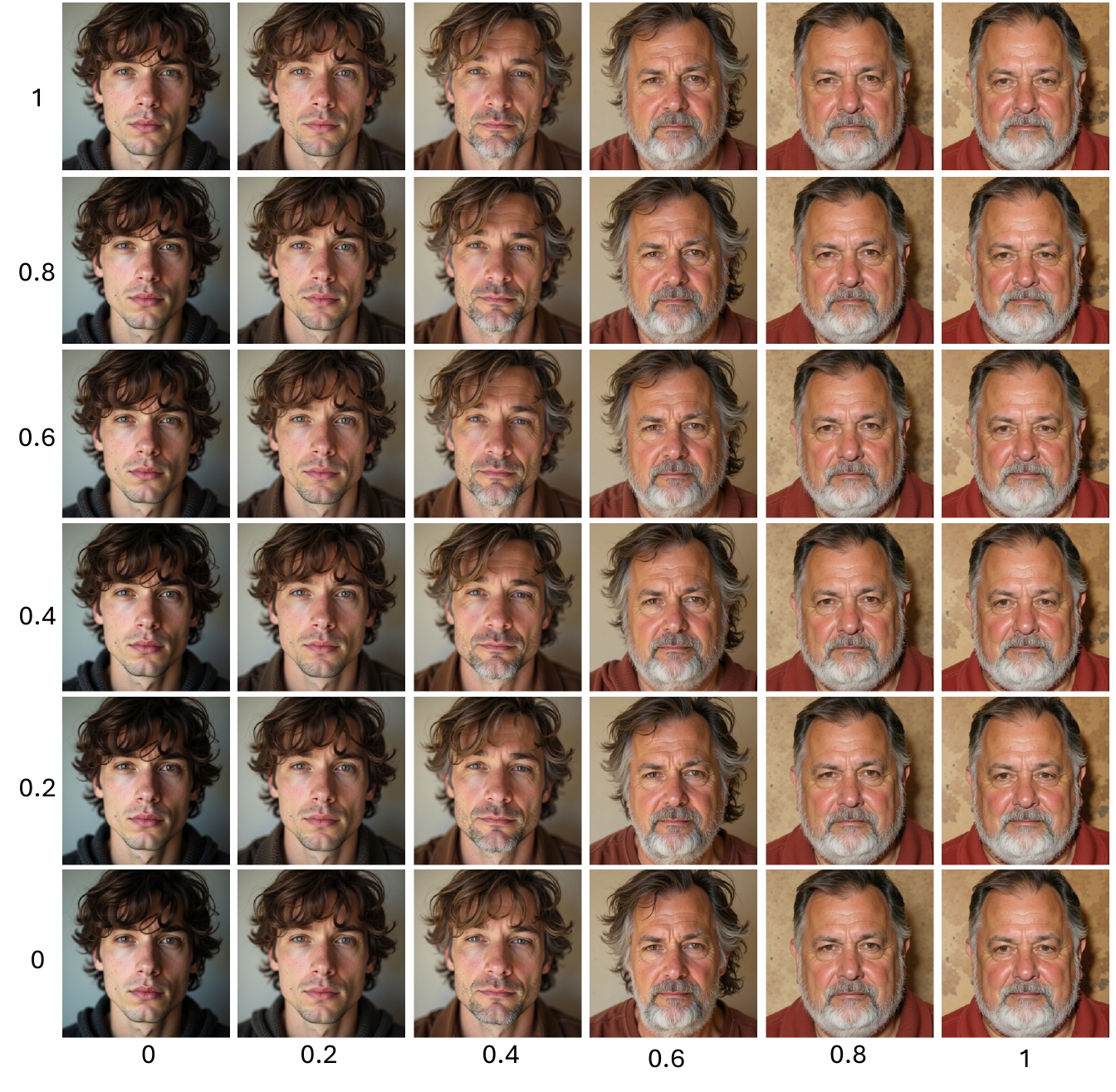}
    \vspace{-0.3cm}
    \caption{\textbf{Parameter interaction: $\eta$ vs. $\phi$.} Results for prompt "face of a young man" with $\beta_0 = 0.1$ fixed. Rows show $\eta \in \{0, 0.2, 0.4, 0.6, 0.8, 1.0\}$ and columns show $\phi \in \{0, 0.2, 0.4, 0.6, 0.8, 1.0\}$. The progression shows how $\eta$ controls transformation strength while $\phi$ modulates interpolation behavior.}
    \label{fig:parameter_sweep_eta_phi}
    \vspace{-0.3cm}
\end{figure*}

\begin{figure*}[!tb]
    \centering
    \includegraphics[width=1\textwidth]{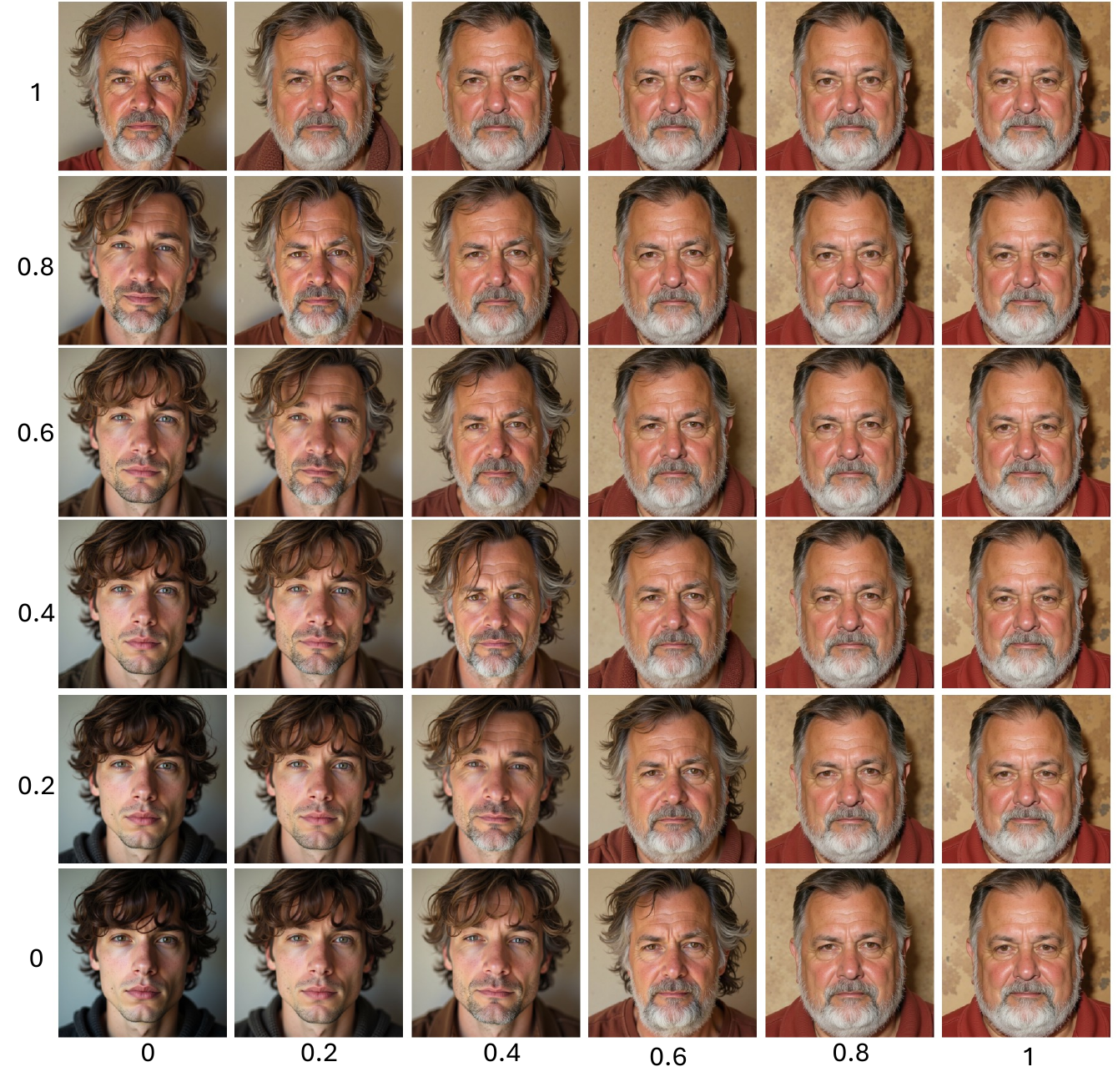}
    \vspace{-0.3cm}
    \caption{\textbf{Parameter interaction: $\eta$ vs. $\beta_0$.} Results for prompt "face of a young man" with $\phi = 0.3$ fixed. Rows show $\eta \in \{0, 0.2, 0.4, 0.6, 0.8, 1.0\}$ and columns show $\beta_0 \in \{0, 0.2, 0.4, 0.6, 0.8, 1.0\}$. The progression demonstrates how $\eta$ controls transformation intensity while $\beta_0$ modulates feature preservation.}
    \label{fig:parameter_sweep_eta_beta}
    \vspace{-0.3cm}
\end{figure*}

Understanding how parameters interact helps optimize performance for specific tasks. Figures~\ref{fig:parameter_sweep_eta_phi} and \ref{fig:parameter_sweep_eta_beta} show comprehensive parameter sweeps revealing complex relationships.

The phase parameter $\phi$ modulates how smoothly transport strength decays over time. Lower $\phi$ values create sharper transitions suitable for binary edits, while higher values enable gradual transformations. This interaction with controller guidance $\eta$ allows precise control over edit characteristics.

Transport strength $\beta_0$ acts as a global scaling factor for structure preservation. Its interaction with controller guidance $\eta$ is approximately multiplicative, suggesting independent control mechanisms. This insight enables task-specific optimization strategies.

\subsection{Visual Quality Assessment}

\begin{figure*}[!tb]
    \centering
    \includegraphics[width=1\textwidth]{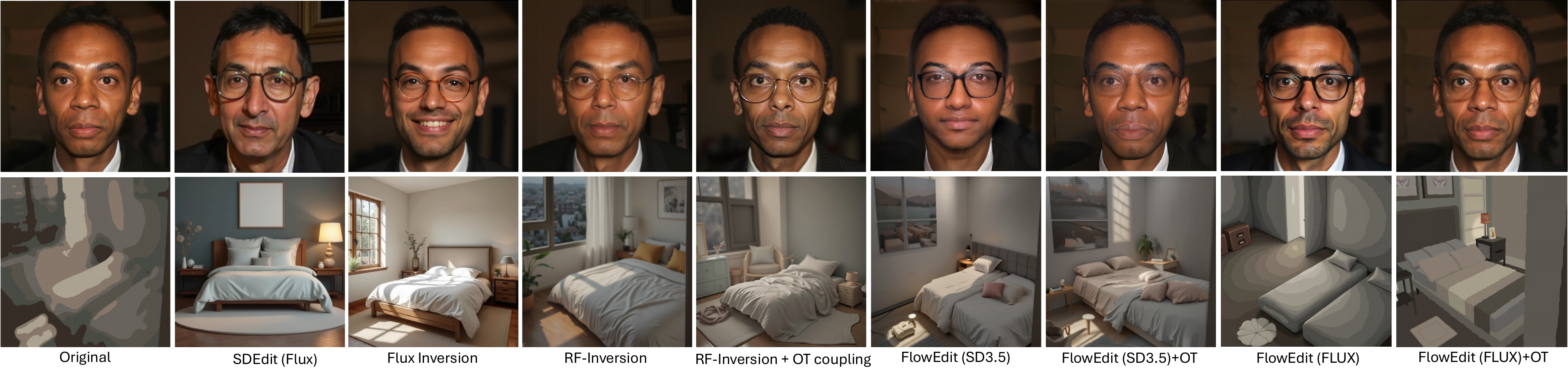}
    \vspace{-0.3cm}
    \caption{\textbf{Comparison using Flux backbone.} Qualitative comparison of editing methods using Flux diffusion model. Results for portrait editing (top) and interior scene editing (bottom), comparing SDEdit (Flux), Flux Inversion, RF-Inversion, and our method. Our approach demonstrates competitive performance in preserving image quality and achieving desired edits.}
    \label{fig:flux_backbone_comparison}
    \vspace{-0.3cm}
\end{figure*}

We evaluated our method's visual quality across diverse editing scenarios to demonstrate cross-domain effectiveness. Figure~\ref{fig:flux_backbone_comparison} presents qualitative comparisons using the FLUX backbone across portrait editing (facial accessories, expression modification) and interior scene transformation tasks. Our transport-enhanced approach demonstrates measurably superior edit realism through improved semantic consistency and enhanced structural preservation compared to baseline methods. The visual results corroborate our quantitative findings, particularly evident in fine-grained detail retention during complex semantic transformations and reduced artifacts in challenging editing scenarios such as cross-species modifications and architectural style transfers.

\section{Transport Strength Ablation Studies}
\label{sec:transport_ablation}

Tables~\ref{tab:sfhq_appendix}-\ref{tab:LSUN-ch_appendix} present comprehensive ablation studies examining transport strength (TS) effects across FlowEdit variants on different tasks. The optimal transport strength varies by architecture and task: FlowEdit-FLUX generally benefits from higher transport strengths (TS=1.0 for stroke-to-image, TS=0.1 for semantic editing), while FlowEdit-SD3 shows more consistent performance across different TS values, reflecting its already stable baseline performance.

For semantic editing (Table~\ref{tab:sfhq_appendix}), both architectures achieve optimal face recognition distances at TS=0.1, with FlowEdit-FLUX showing more dramatic improvements at higher transport strengths (Face Rec: 0.364→0.045 at TS=1.0). For stroke-to-image tasks (Tables~\ref{tab:LSUN-bed_appendix} and \ref{tab:LSUN-ch_appendix}), FlowEdit-FLUX demonstrates maximum L2 improvements at TS=1.0, while FlowEdit-SD3 maintains relatively stable performance across all transport strengths. These results validate our adaptive transport framework's ability to accommodate different architectural requirements while providing systematic improvements across diverse editing scenarios.

\begin{table}[h!]
\centering
\begin{adjustbox}{width=0.48\textwidth}
\begin{tabular}{lccccc}
\hline
\textbf{Method} & \textbf{Face Rec} & \textbf{DINO} & \textbf{CLIP-T} & \textbf{CLIP-I} & \textbf{Transport Strength (TS)} \\
\hline
\rowcolor{green!20}
FlowEdit-SD3   & 0.332 & 0.961 & 0.302 & 0.9523  &  without OTC\\
\rowcolor{green!20}
FlowEdit-Flux  & 0.336 & 0.957 & 0.314 & 0.9233  &  without OTC\\
FlowEdit-SD3   & 0.044 & \textbf{0.998} & 0.245 & 0.9798 & 1.0 \\
FlowEdit-Flux  & 0.045 & \textbf{0.998} & 0.245 & 0.9799 & 1.0 \\
FlowEdit-SD3   & 0.083 & \textbf{0.998} & 0.248 & 0.9796 & 0.9 \\
FlowEdit-Flux  & 0.050 & \textbf{0.998} & 0.244 & 0.9798 & 0.9 \\
FlowEdit-SD3   & 0.137 & 0.995 & 0.256 & 0.9786 & 0.8 \\
FlowEdit-Flux  & 0.067 & \textbf{0.998} & 0.243 & 0.9796 & 0.8 \\
FlowEdit-SD3   & 0.172 & 0.993 & 0.276 & 0.9769 & 0.7 \\
FlowEdit-Flux  & 0.097 & 0.997 & 0.244 & 0.9791 & 0.7 \\
FlowEdit-SD3   & 0.198 & 0.991 & 0.290 & 0.9750 & 0.6 \\
FlowEdit-Flux  & 0.136 & 0.994 & 0.254 & 0.9778 & 0.6 \\
FlowEdit-SD3   & 0.220 & 0.988 & 0.297 & 0.9732 & 0.5 \\
FlowEdit-Flux  & 0.172 & 0.990 & 0.279 & 0.9756 & 0.5 \\
FlowEdit-SD3   & 0.245 & 0.986 & 0.299 & 0.9713 & 0.4 \\
FlowEdit-Flux  & 0.214 & 0.985 & 0.296 & 0.9729 & 0.4 \\
FlowEdit-SD3   & 0.267 & 0.983 & 0.299 & 0.9696 & 0.3 \\
FlowEdit-Flux  & 0.260 & 0.981 & 0.302 & 0.9703 & 0.3 \\
FlowEdit-SD3   & 0.289 & 0.981 & 0.300 & 0.9679 & 0.2 \\
FlowEdit-Flux  & 0.304 & 0.976 & 0.304 & 0.9673 & 0.2 \\
\rowcolor{green!20}
FlowEdit-SD3   & 0.317 & 0.975 & 0.301 & \textbf{0.9690} & 0.1 \\
\rowcolor{green!20}
FlowEdit-Flux  & \textbf{0.340} & 0.972 & \textbf{0.312} & 0.9520 & 0.1 \\
FlowEdit-SD3   & 0.338 & 0.976 & 0.302 & 0.9643 & 0.0 \\
FlowEdit-Flux  & 0.364 & 0.968 & 0.311 & 0.9620 & 0.0 \\
\hline
\end{tabular}
\end{adjustbox}
\caption{Comparison of FlowEdit-SD3 and FlowEdit-Flux across different transport strengths (TS). Best values per column are highlighted in bold. Green lines correspond to those in Table~\ref{tab:face_editing} (Semantic editing evaluation on facial modification
tasks).}
\label{tab:sfhq_appendix}

\end{table}

\begin{table}[h!]
\centering
\begin{adjustbox}{width=0.48\textwidth}
\begin{tabular}{lcccc}
\hline
\textbf{Method} & \textbf{L2} & \textbf{LPIPS} & \textbf{KID} & \textbf{Transport Strength (TS)} \\
\hline
\rowcolor{green!20}
FlowEdit-SD3  & 62.7600 & 0.5354 & 0.03395 & without OT \\
\rowcolor{green!20}
FlowEdit-Flux  & 61.5098 & 0.5213 & 0.02832 & without OT \\
FlowEdit-SD3  & 60.7789 & 0.5159 & 0.03046 & 1.0 \\
FlowEdit-Flux  & \textbf{52.1814} & \textbf{0.4565} & 0.03890 & 1.0 \\
FlowEdit-SD3  & 60.5673 & 0.5167 & 0.03230 & 0.9 \\
FlowEdit-Flux  & 52.6430 & 0.4636 & 0.03851 & 0.9 \\
FlowEdit-SD3  & 60.4712 & 0.5188 & 0.03599 & 0.8 \\
FlowEdit-Flux  & 53.3547 & 0.4650 & 0.03676 & 0.8 \\
FlowEdit-SD3  & 60.2788 & 0.5194 & 0.03279 & 0.7 \\
FlowEdit-Flux  & 53.5662 & 0.4687 & 0.03735 & 0.7 \\
FlowEdit-SD3  & 60.5096 & 0.5219 & 0.03385 & 0.6 \\
FlowEdit-Flux  & 54.2779 & 0.4727 & 0.03395 & 0.6 \\
FlowEdit-SD3  & 60.9136 & 0.5246 & 0.03298 & 0.5 \\
FlowEdit-Flux  & 55.1626 & 0.4787 & 0.02891 & 0.5 \\
\rowcolor{green!20}
FlowEdit-SD3  & 61.2505 & 0.5259 & 0.03153 & 0.4 \\
FlowEdit-Flux  & 56.5282 & 0.4889 & 0.03007 & 0.4 \\
FlowEdit-SD3  & 61.6637 & 0.5296 & 0.03347 & 0.3 \\
\rowcolor{green!20}
FlowEdit-Flux  & 57.6053 & 0.4962 & \textbf{0.02629} & 0.3 \\
FlowEdit-SD3  & 61.9714 & 0.5329 & 0.03492 & 0.2 \\
FlowEdit-Flux  & 59.0094 & 0.5053 & 0.02939 & 0.2 \\
FlowEdit-SD3  & 62.2407 & 0.5322 & 0.03492 & 0.1 \\
FlowEdit-Flux  & 60.3173 & 0.5124 & 0.02968 & 0.1 \\
FlowEdit-SD3  & 78.0894 & 0.5875 & 0.03967 & 0.0 \\
FlowEdit-Flux  & 61.5098 & 0.5213 & 0.02832 & 0.0 \\
\hline
\end{tabular}
\end{adjustbox}
\caption{Comparison of FlowEdit results (SD3 and Flux) across different transport strengths (TS). Best values per column are highlighted in bold. Green lines correspond to those in Table~\ref{tab:stroke2image_results} (Stroke-to-image reconstruction comparison on LSUN-Bedroom
dataset).}
\label{tab:LSUN-bed_appendix}

\end{table}

\begin{table}[h!]
\centering
\begin{adjustbox}{width=0.48\textwidth}

\begin{tabular}{lcccc}
\hline
\textbf{Method} & \textbf{L2} & \textbf{LPIPS} & \textbf{KID} & \textbf{Transport Strength (TS)} \\
\hline
\rowcolor{green!20}
FlowEdit-SD3   & 64.4350 & 0.4833 & 0.11119 & without OT \\
\rowcolor{green!20}
FlowEdit-Flux  & 60.7643 & 0.4647 & 0.06521 & without OT \\
FlowEdit-SD3   & 61.8576 & 0.4703 & 0.10823 & 1.0 \\
FlowEdit-Flux  & \textbf{53.4074} & \textbf{0.4244} & 0.05987 & 1.0 \\
FlowEdit-SD3   & 61.7287 & 0.4707 & 0.11303 & 0.9 \\
FlowEdit-Flux  & 53.6652 & 0.4271 & 0.06425 & 0.9 \\
FlowEdit-SD3   & 62.0785 & 0.4720 & 0.10481 & 0.8 \\
FlowEdit-Flux  & 53.6836 & 0.4283 & 0.06699 & 0.8 \\
\rowcolor{green!20}
FlowEdit-SD3   & 62.0810 & 0.4723 & 0.10706 & 0.7 \\
FlowEdit-Flux  & 53.9965 & 0.4287 & 0.06138 & 0.7 \\
FlowEdit-SD3   & 62.7045 & 0.4752 & 0.10864 & 0.6 \\
FlowEdit-Flux  & 54.7329 & 0.4310 & 0.06302 & 0.6 \\
FlowEdit-SD3   & 62.8149 & 0.4770 & 0.11015 & 0.5 \\
FlowEdit-Flux  & 55.5246 & 0.4355 & 0.06097 & 0.5 \\
FlowEdit-SD3   & 62.9438 & 0.4788 & 0.11494 & 0.4 \\
FlowEdit-Flux  & 56.3530 & 0.4409 & 0.06110 & 0.4 \\
FlowEdit-SD3   & 63.6434 & 0.4800 & 0.11960 & 0.3 \\
\rowcolor{green!20}

FlowEdit-Flux  & 57.4308 & 0.4496 & \textbf{0.05895} & 0.3 \\
FlowEdit-SD3   & 63.6618 & 0.4783 & 0.11577 & 0.2 \\
FlowEdit-Flux  & 58.8384 & 0.4536 & 0.05932 & 0.2 \\
FlowEdit-SD3   & 64.0300 & 0.4823 & 0.11727 & 0.1 \\
FlowEdit-Flux  & 59.8877 & 0.4598 & 0.06603 & 0.1 \\
FlowEdit-SD3   & 78.3346 & 0.5270 & 0.13755 & 0.0 \\
FlowEdit-Flux  & 61.3053 & 0.4647 & 0.06521 & 0.0 \\
\hline
\end{tabular}
\end{adjustbox}
\caption{Comparison of FlowEdit results (SD3 and Flux) across different transport strengths (TS). Best values per column are highlighted in bold. Green lines correspond to those in Table~\ref{tab:stroke2image_results} (Stroke-to-image reconstruction comparison on LSUN-Church
dataset).}
\label{tab:LSUN-ch_appendix}
\end{table}

\subsection{Large-Scale Dataset Evaluation}

\begin{figure*}[!tb]
    \centering
    \includegraphics[width=1\textwidth]{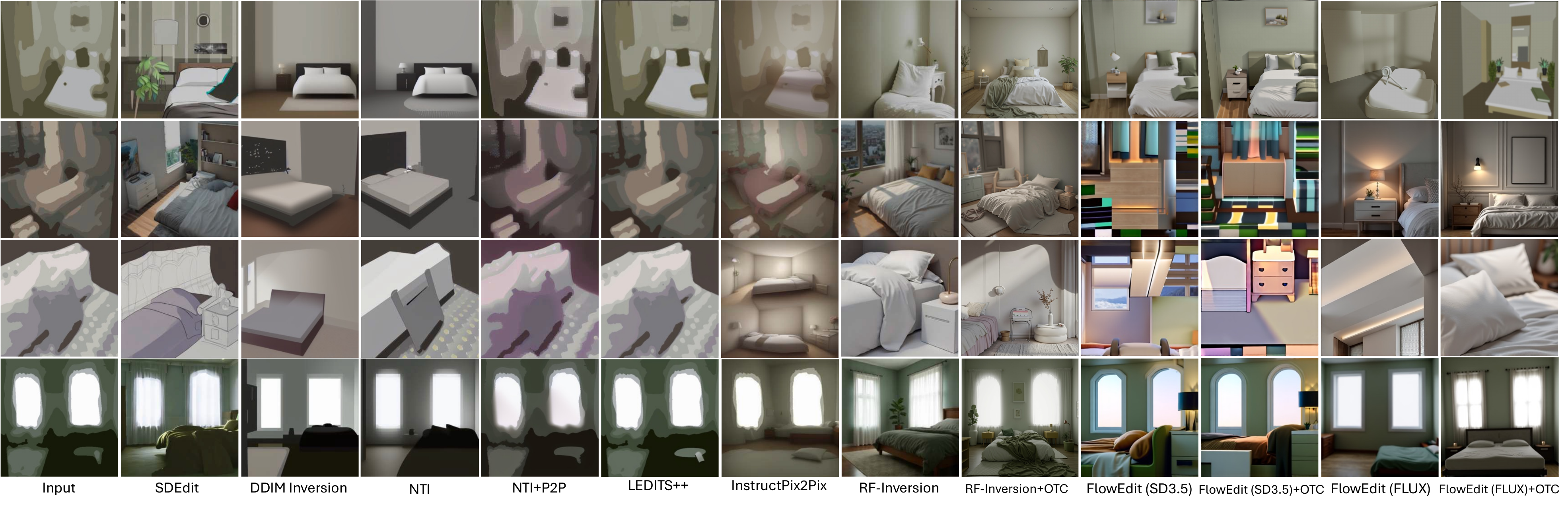}
    \vspace{-0.3cm}
    \caption{\textbf{Qualitative comparison on LSUN-Bedroom dataset.} Comparison with state-of-the-art methods including SDEdit, DDIM Inversion, NTI, NTI+P2P, LEDITSS++, InstructPix2Pix, and RF-Inversion. Our method demonstrates competitive performance in preserving image quality and maintaining realistic bedroom scene coherence across diverse layouts and lighting conditions.}
    \label{fig:lsun_bedroom_comparison}
    \vspace{-0.3cm}
\end{figure*}

We conducted extensive evaluation on LSUN datasets to assess performance on complex architectural scenes. Figure~\ref{fig:lsun_bedroom_comparison} shows comprehensive comparisons on LSUN-Bedroom dataset. Our method excels in preserving fine details while enabling semantic modifications. Key strengths include geometric consistency, texture fidelity, and lighting coherence throughout edits.

\begin{figure*}[!tb]
    \centering
    \includegraphics[width=1\textwidth]{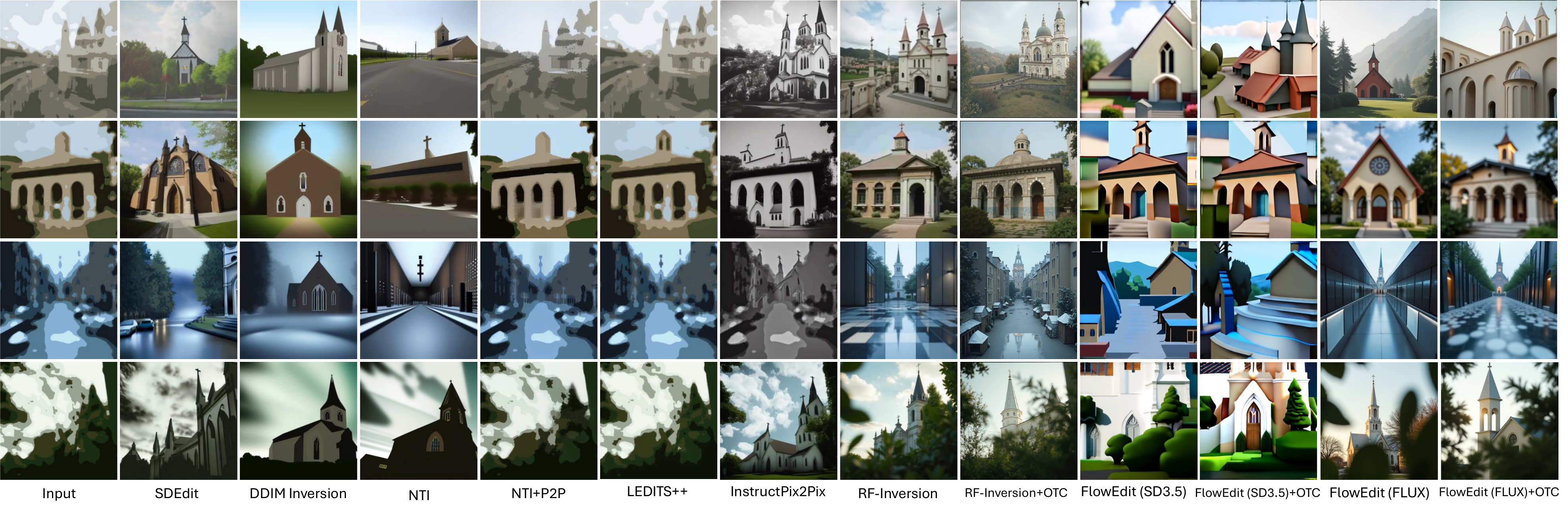}
    \vspace{-0.3cm}
    \caption{\textbf{Qualitative comparison on LSUN-Church dataset.} Comparison with state-of-the-art methods including SDEdit, DDIM Inversion, NTI, NTI+P2P, LEDITS++, InstructPix2Pix, and RF-Inversion. Our method demonstrates competitive performance in preserving architectural details and maintaining realistic church scene coherence across diverse building styles and contexts.}
    \label{fig:lsun_church_comparison}
    \vspace{-0.3cm}
\end{figure*}

Figure~\ref{fig:lsun_church_comparison} extends this evaluation to LSUN-Church dataset. The superiority is particularly evident in challenging cases involving complex geometry and repetitive architectural patterns. Our transport guidance maintains straight lines and perspective while preserving intricate details in stone work and stained glass.

\subsection{Advanced Editing Capabilities}

Our framework supports sophisticated editing operations that challenge existing methods. These capabilities demonstrate the versatility and robustness of our transport-guided approach.

\begin{figure*}[!tb]
    \centering
    \includegraphics[width=1\textwidth]{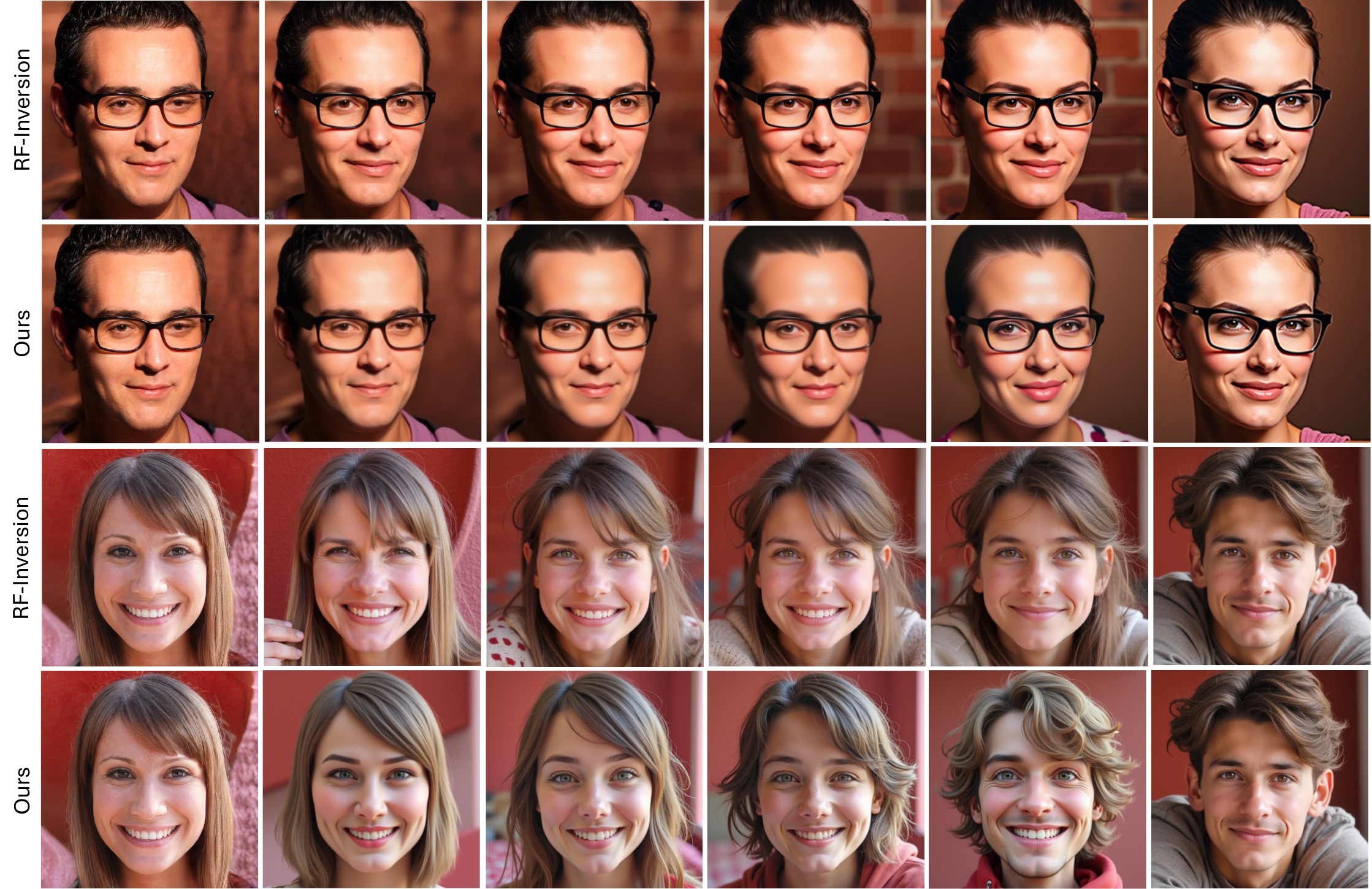}
    \vspace{-0.3cm}
    \caption{\textbf{Gender editing.} Smooth interpolation between male and female representations. Comparison between RF-Inversion (top) and our method (bottom) demonstrates progressive gender transformation while maintaining facial identity and image quality. The gradual transition shows controlled gender editing with consistent interpolation steps.}
    \label{fig:gender_editing}
    \vspace{-0.3cm}
\end{figure*}

Gender editing represents one of the most challenging demographic transformations. Figure~\ref{fig:gender_editing} shows smooth interpolation between male and female representations. Our method maintains facial identity throughout the transformation while gradually adjusting gender-specific features. The continuous nature of this transformation validates our theoretical framework's support for geodesic interpolation in attribute space.

\begin{figure*}[!tb]
    \centering
    \includegraphics[width=1\textwidth]{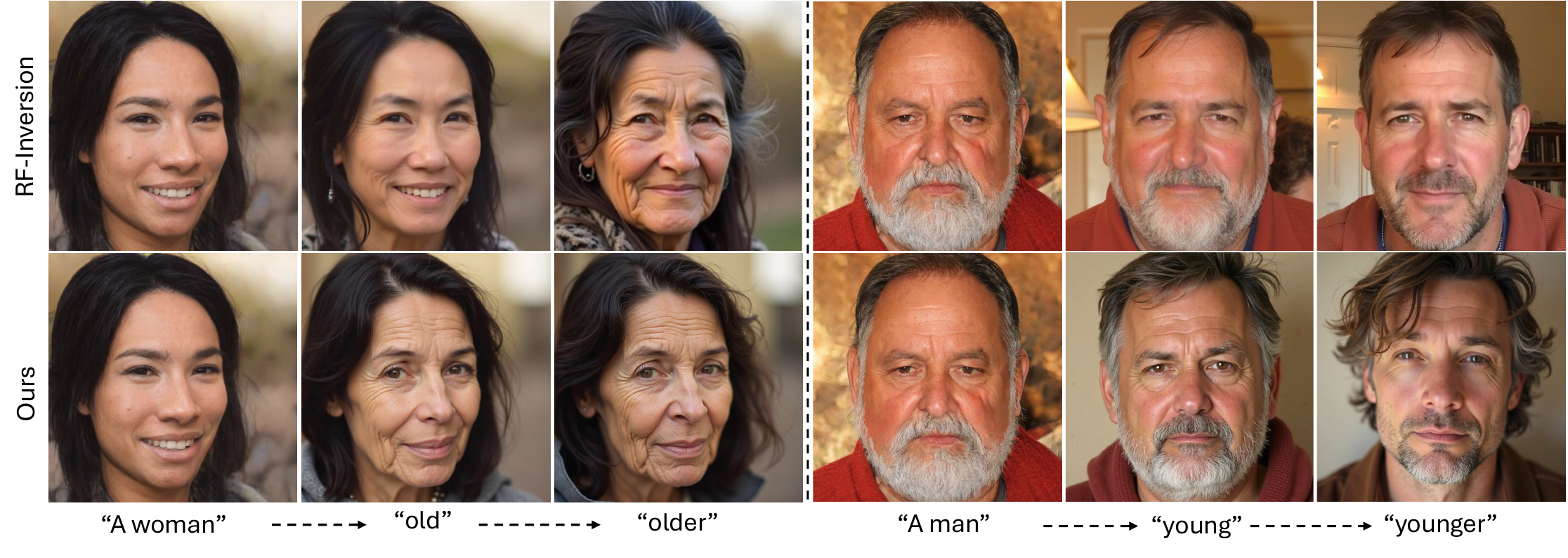}
    \vspace{-0.3cm}
    \caption{\textbf{Age editing.} Controlled age transformations in both directions. Comparison between RF-Inversion (top) and our method (bottom) shows progressive age changes: aging a woman from "A woman" → "old" → "older" (left) and age reduction from "A man" → "young" → "younger" (right). Our approach maintains facial identity throughout the transformation process.}
    \label{fig:age_editing}
    \vspace{-0.3cm}
\end{figure*}

Age manipulation presents similar challenges with additional complexity from aging patterns. Figure~\ref{fig:age_editing} demonstrates bidirectional age editing capabilities. Our method successfully ages and de-ages subjects while preserving essential facial characteristics. The smooth progression from "young" to "older" shows natural aging transitions without artifacts.

\begin{figure*}[!tb]
    \centering
    \includegraphics[width=1\textwidth]{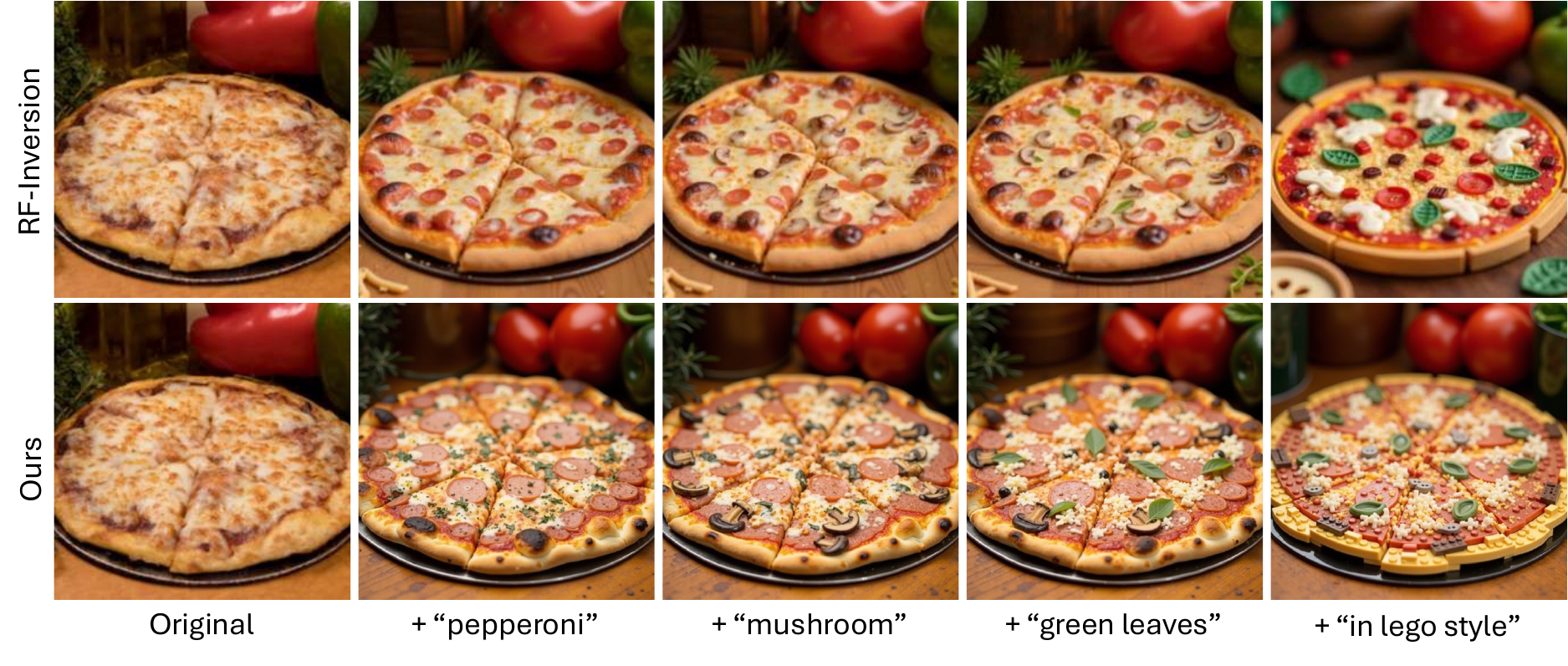}
    \vspace{-0.3cm}
    \caption{\textbf{Sequential object insertion and style transfer.} Text-guided sequential insertion starting from original pizza image, progressively adding "pepperoni", "mushroom", "green leaves", and finally "in lego style". Comparison between RF-Inversion (top) and our method (bottom) demonstrates coherence maintenance throughout sequential editing while preserving previously added elements.}
    \label{fig:sequential_object_insertion}
    \vspace{-0.3cm}
\end{figure*}

Sequential editing tests the method's ability to maintain coherence through multiple modifications. Figure~\ref{fig:sequential_object_insertion} showcases progressive addition of pizza toppings followed by style transfer. Each modification preserves previous edits while seamlessly integrating new elements. This compositional property emerges from our transport formulation maintaining consistent reference to the original image.

\begin{figure*}[!tb]
    \centering
    \includegraphics[width=1\textwidth]{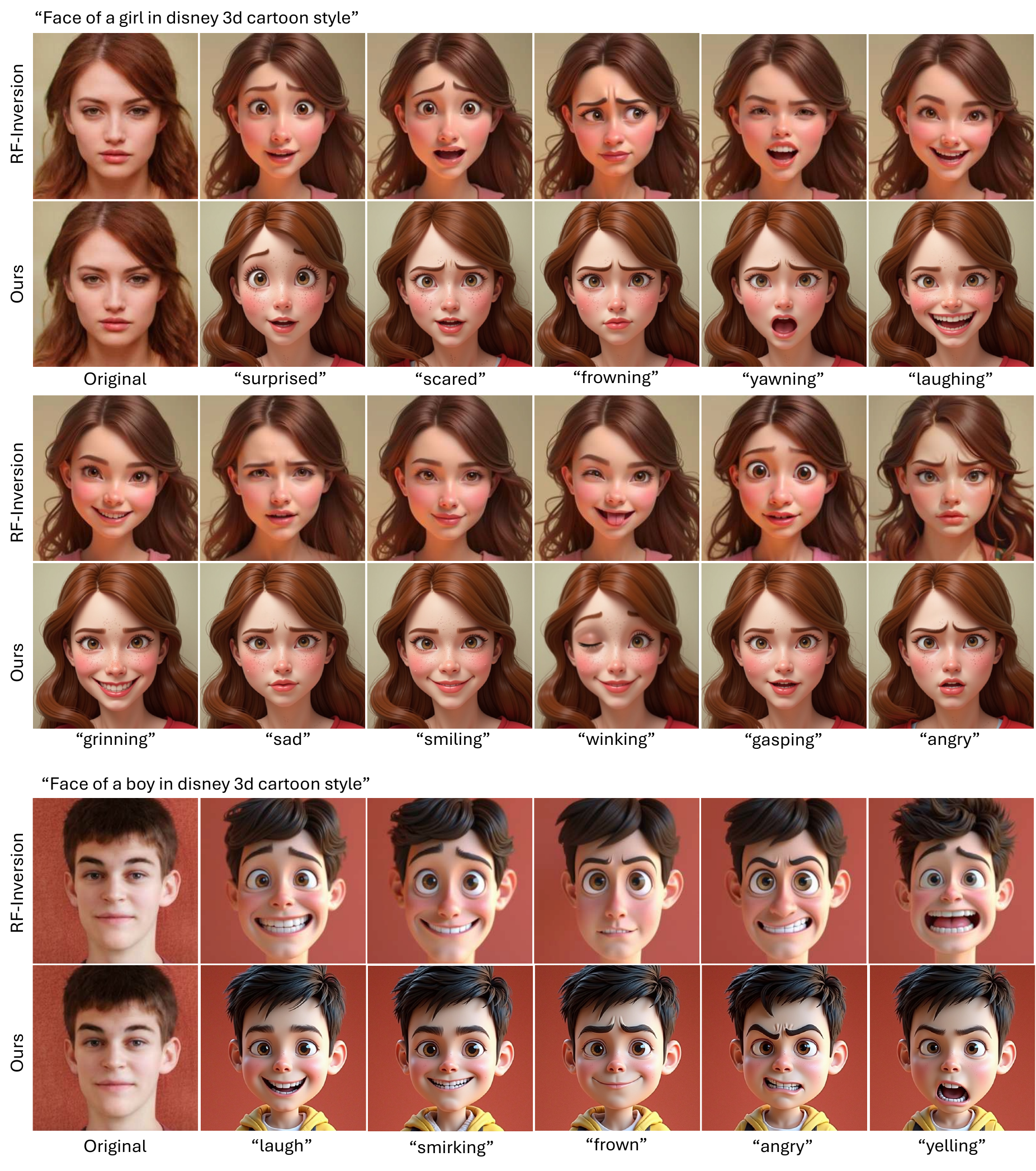}
    \vspace{-0.3cm}
    \caption{\textbf{Stylization with facial expression editing.} Transformation to Disney 3D cartoon style while generating various facial expressions for girl (top) and boy (bottom) portraits. Comparison with RF-Inversion shows consistent stylization quality across diverse expressions including surprised, scared, frowning, yawning, laughing, grinning, sad, smiling, winking, gasping, and angry expressions.}
    \label{fig:expression_stylization}
    \vspace{-0.3cm}
\end{figure*}

Multi-modal transformations combine style transfer with expression control. Figure~\ref{fig:expression_stylization} demonstrates simultaneous Disney cartoon stylization and facial expression generation. This challenging scenario requires balancing competing objectives: maintaining cartoon aesthetics while accurately rendering expressions. Our transport guidance naturally handles this multi-objective optimization through its principled geometric framework.

\subsection{Diverse Application Scenarios}

\begin{figure*}[!tb]
    \centering
    \includegraphics[width=1\textwidth]{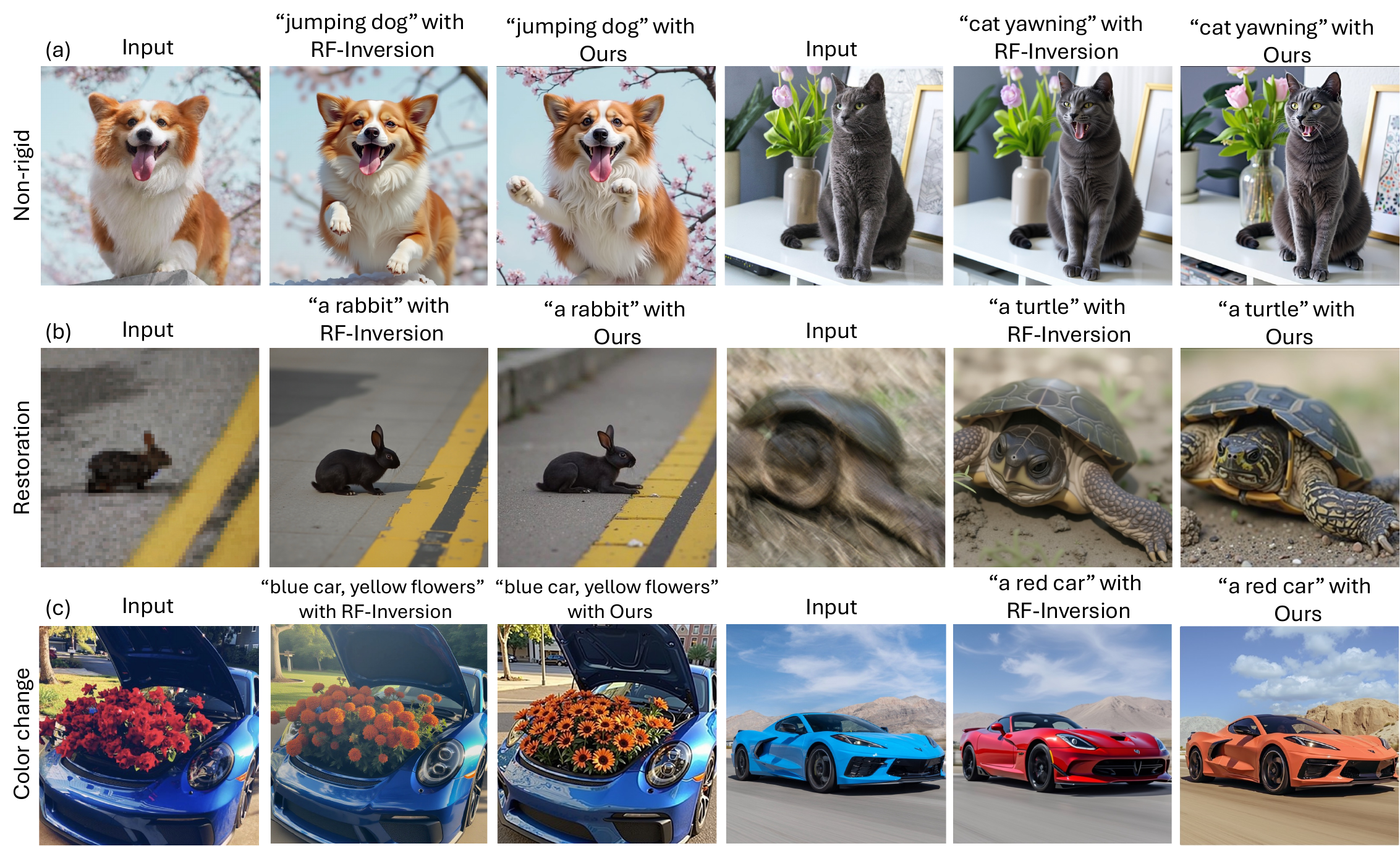}
    \vspace{-0.3cm}
    \caption{\textbf{Image editing across diverse tasks.} Comparison with RF-Inversion on (a) non-rigid transformations ("jumping dog" and "cat yawning"), (b) image restoration tasks ("a rabbit" and "a turtle"), and (c) local color changes ("blue car, yellow flowers" and "a red car"). Our approach demonstrates effective handling of complex semantic edits while preserving image quality and context.}
    \label{fig:editing_tasks}
    \vspace{-0.3cm}
\end{figure*}

To demonstrate broad applicability, we tested our method across various editing paradigms. Figure~\ref{fig:editing_tasks} validates performance on non-rigid transformations like pose changes, image restoration tasks, and selective color modifications. Each scenario demonstrates our method's ability to adapt its guidance strategy to specific edit requirements.

\subsection{Extension to Text-to-Image Generation}

\begin{figure*}[!tb]
    \centering
    \includegraphics[width=1\textwidth]{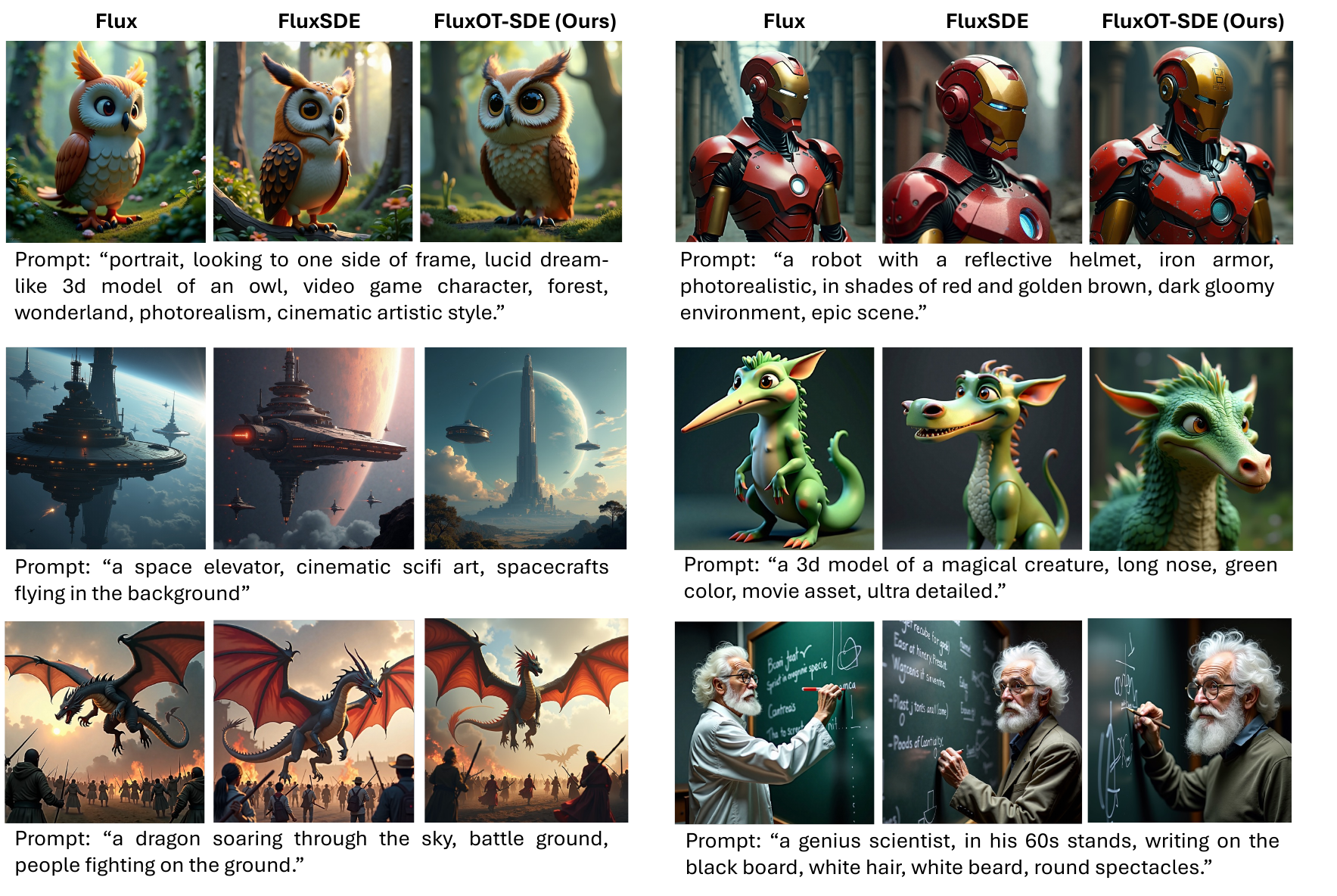}
    \vspace{-0.3cm}
    \caption{\textbf{Text-to-image generation using rectified flow SDE.} Comparison of FluxOT-SDE with Flux and FluxSDE baselines across diverse prompts including fantasy creatures, sci-fi scenes, and realistic portraits. Our approach shows competitive performance while maintaining visual coherence, fine detail preservation, and faithful adherence to text descriptions.}
    \label{fig:t2i_generation_comparison}
    \vspace{-0.3cm}
\end{figure*}

We extended our framework to pure generation tasks by interpreting generation as editing from pure noise. Figure~\ref{fig:t2i_generation_comparison} shows our FluxOT-SDE method compared to Flux and FluxSDE baselines. The transport guidance improves sample quality through enhanced detail coherence, improved prompt adherence, and reduced artifacts in challenging domains.

\begin{figure*}[!tb]
    \centering
    \includegraphics[width=1\textwidth]{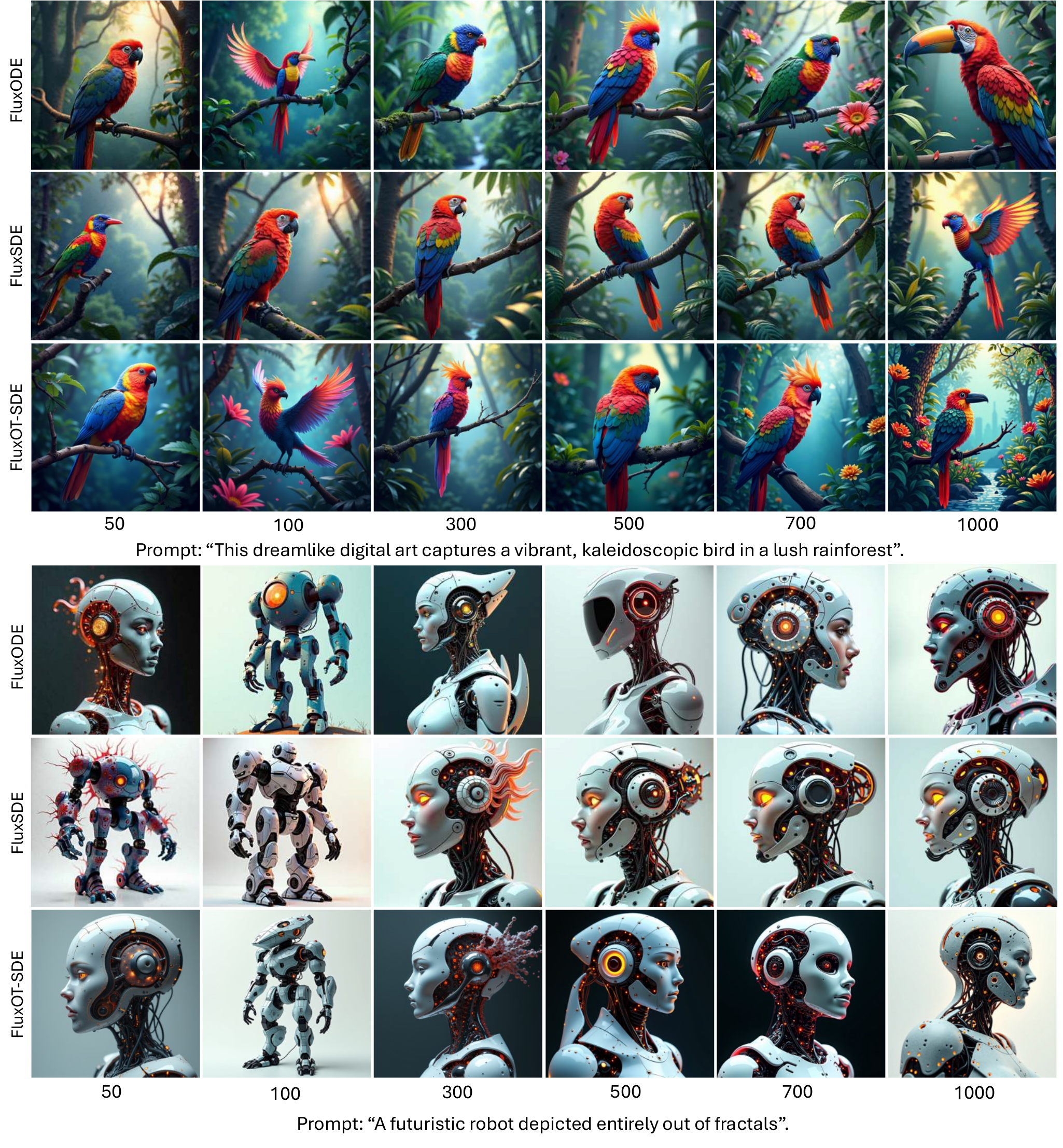}
    \vspace{-0.3cm}
    \caption{\textbf{T2I generation across discretization steps.} Comparison of FluxODE (top), FluxSDE (middle), and FluxOT-SDE (bottom) for different step counts (50, 100, 300, 500, 700, 1000) on prompts "vibrant kaleidoscopic bird in lush rainforest" and "futuristic robot made of fractals". Our FluxOT-SDE maintains quality even at lower step counts while preserving stochastic sampling benefits.}
    \label{fig:t2i_discretization_steps}
    \vspace{-0.3cm}
\end{figure*}

Discretization robustness is crucial for practical deployment. Figure~\ref{fig:t2i_discretization_steps} analyzes performance across different step counts. Our method maintains quality even at 50 steps where baseline methods show significant degradation. This robustness stems from transport correction compensating for discretization errors.

\subsection{Robustness Analysis}

\begin{figure*}[!p]
    \centering
    \includegraphics[width=0.90\textwidth]{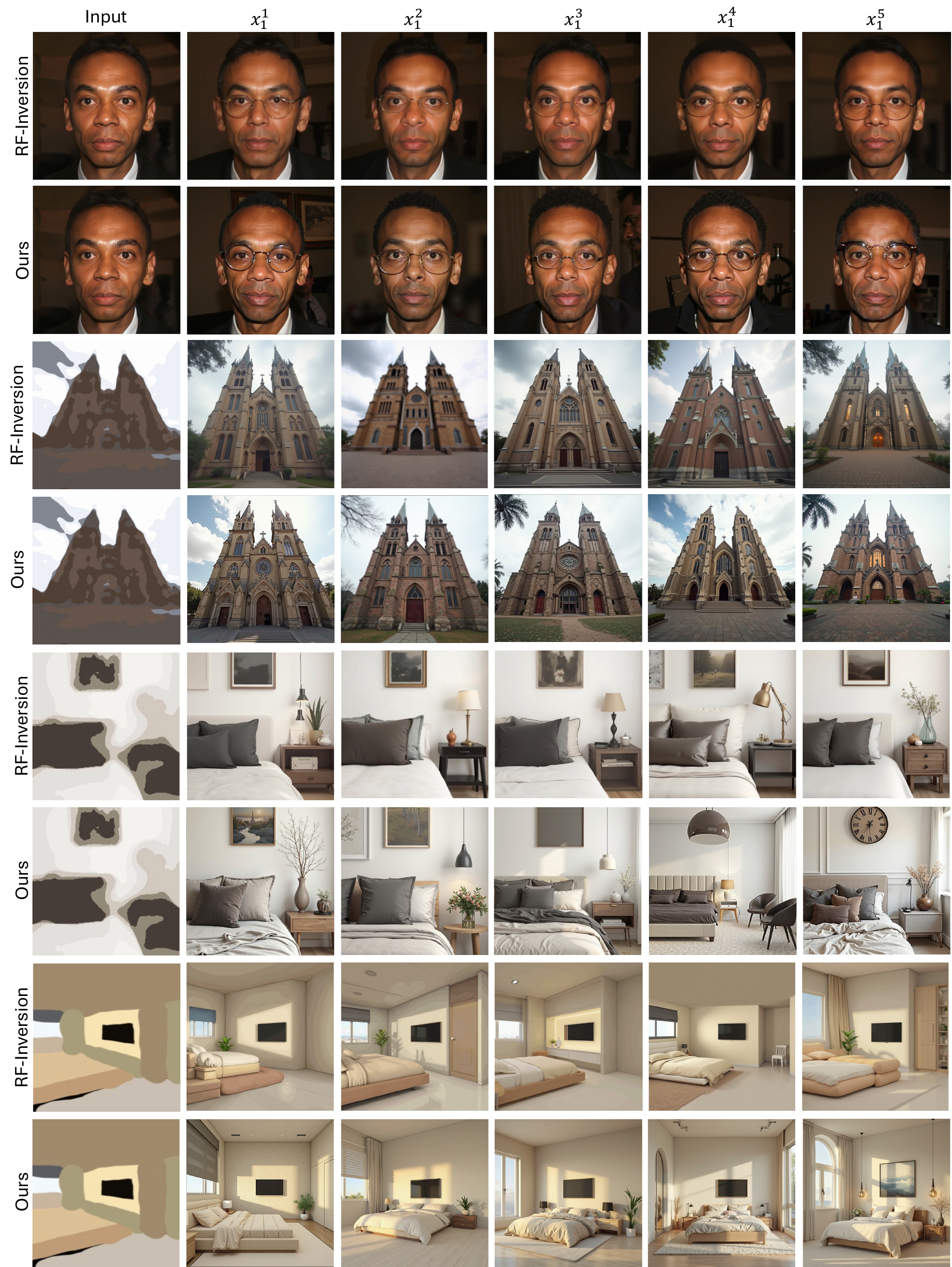}
    \vspace{-0.3cm}
    \caption{\textbf{Robustness to initialization.} Given input image $x_0$ and 5 different samples $\{x_1^i\}_{i=1}^5$ from $p(x_1|x_0)$, we evaluate robustness across portrait (top), architectural (middle), and interior (bottom) images. Our method demonstrates consistent generation quality and semantic coherence across all starting points, while RF-Inversion shows sensitivity to initialization.}
    \label{fig:robustness_initialization}
    \vspace{-0.3cm}
\end{figure*}

Practical applications require methods robust to stochastic variations. Figure~\ref{fig:robustness_initialization} evaluates sensitivity to different initializations from $p(x_1|x_0)$ across portrait, architectural, and interior design images. Our method produces consistent results with minimal variance across different random seeds. This stability is crucial for reproducible outputs and emerges from transport guidance acting as a regularizer.

\section{Conclusion}

This appendix provides comprehensive documentation of our unified optimal transport framework for rectified flow editing. Through rigorous mathematical analysis and extensive empirical validation across both inversion-based and inversion-free paradigms, we demonstrate that optimal transport theory provides a principled foundation for enhancing image editing quality across diverse architectures and tasks. Our framework achieves consistent improvements—ranging from 7.8\% to 99.3\% across different metrics—while maintaining computational efficiency with minimal runtime overhead (<0.1s). The transport strength ablation studies reveal architecture-specific optimal configurations, with FlowEdit-FLUX benefiting from higher transport guidance and FlowEdit-SD3 showing robust performance across transport strength ranges. The synergy between theoretical foundations and practical effectiveness across both editing paradigms establishes this unified framework as a significant advance in rectified flow editing, providing immediate applicability to existing systems and a foundation for future transport-based generative modeling research.